\documentclass[a4paper,fleqn]{cas-dc}

\usepackage[numbers]{natbib}
\usepackage{hyperref}
\usepackage{xurl}

\def\tsc#1{\csdef{#1}{\textsc{\lowercase{#1}}\xspace}}
\tsc{WGM}
\tsc{QE}
\tsc{EP}
\tsc{PMS}
\tsc{BEC}
\tsc{DE}

\usepackage{amsmath}
\usepackage{amssymb}
\usepackage{booktabs}
\usepackage{amsthm}

\usepackage{placeins}
\usepackage{nicefrac}
\usepackage{algorithm}
\usepackage{algpseudocode}
\usepackage{array}
\usepackage[inline]{enumitem}
\usepackage{multirow}
\usepackage{overpic}
\usepackage{multirow, makecell}
\usepackage{bm}
\usepackage{wrapfig}
\usepackage{xspace}
\usepackage{subcaption}
\usepackage{placeins}
\usepackage{cleveref}

\def\etal{\textit{et al.}\xspace}

\renewcommand{\vec}{\mathbf}
\newcommand{\ransac}{\textsc{RanSaC}\xspace}
\newcommand{\ransacov}{\textsc{RanSaCov}\xspace}
\newcommand{\gcransac}{\textsc{GC-RanSaC}\xspace}
\newcommand{\loransac}{\textsc{LO-RanSaC}\xspace}

\begin{document}
\let\WriteBookmarks\relax
\def\floatpagepagefraction{1}
\def\textpagefraction{.001}
\shorttitle{Superquadric Primitive Decomposition}
\shortauthors{A. Rinaldi \etal}

\title [mode = title]{Superquadric Primitive Decomposition of 3D point clouds via Geometric-Aware Inlier Refinement}

\author[1]{Alessandro Rinaldi}[orcid=]
\cormark[1]
\ead{}

\author[2]{Edoardo Tedesco}[orcid=0009-0000-5574-6860]
\cormark[1]
\ead{edoardo.tedesco@mail.polimi.it}

\author[2]{Andrea Ferraris}[orcid=0009-0003-5383-9299]
\ead{}

\author[2]{Filippo Leveni}[orcid=0009-0007-7745-5686]
\ead{}

\author[3]{Daniele Baieri}[orcid=0000-0002-0704-5960]
\ead{dbaieri@uni-bonn.de}

\author[4]{Filippo Maggioli}[orcid=0000-0001-8008-8468]
\ead{maggioli.filippo@gmail.com}

\author[1]{Simone Melzi}[orcid=0000-0003-2790-9591]
\ead{simone.melzi@unimib.it}

\author[2]{Luca Magri}[orcid=0000-0002-0598-8279]
\cormark[2]
\ead{luca.magri@polimi.it}

\affiliation[1]{organization={Department of Informatics, System and Communication, University of Milano-Bicocca},
                city={Milan},
                country={Italy}}

\affiliation[2]{organization={Department of Electronics, Information and Bioengineering, Politecnico di Milano},
                city={Milan},
                country={Italy}}

\affiliation[3]{organization={Department of Computer Science, University of Bonn},
                city={Bonn},
                country={Germany}}
                
\affiliation[4]{organization={Department of Information Science and Technology, Pegaso University},
                city={Naples},
                country={Italy}}

\cortext[1]{Equal contributor.}
\cortext[2]{Corresponding author.}

\begin{abstract}
The decomposition of 3D point clouds into interpretable geometric primitives remains a longstanding challenge in Computer Vision and Computer Graphics. Among the available representations, superquadrics offer a compact and expressive model capable of capturing a wide range of shapes. However, their estimation is inherently challenging, as it requires solving a non-linear optimization problem and is particularly sensitive to noise, outliers, and overlapping structures.

While robust estimation methods such as \ransac and its variants achieve strong performance, they rely primarily on spatial proximity and residual-based criteria, often leading to incorrect inlier assignments across adjacent or complex arrangements of primitives.

In this work, we introduce a geometric-aware framework for primitive decomposition that explicitly incorporates local surface properties into the fitting process. 
Specifically, we propose an inlier refinement step formulated as an energy minimization problem and solved via graph-cut optimization. 
Our formulation integrates geometric priors, such as normal consistency, enabling more reliable inlier selection beyond purely residual-based criteria. The approach naturally applies to both single-model estimation and multi-model decomposition.

By leveraging geometric information beyond point-wise residuals, our method reduces erroneous inlier propagation and stabilizes parameter estimation. Experiments on synthetic \textcolor{black}{and real} datasets show consistent improvements in geometric accuracy, robustness to noise and outliers, and convergence efficiency compared to state-of-the-art \ransac-based methods.

\end{abstract}


\maketitle

\section{Introduction}
\label{sec:introduction}

\begin{figure*}
\centering
\includegraphics[width= \linewidth]{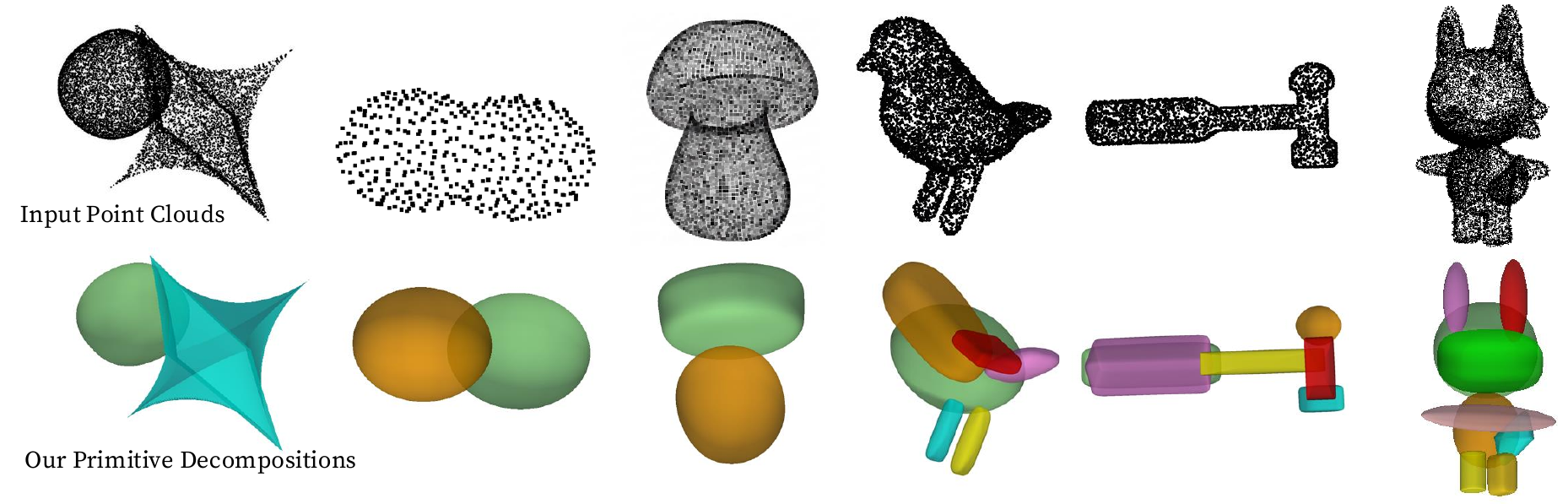}
\caption{Primitive Decomposition of different 3D point clouds (top row) exploiting our Geometric Aware Inlier Refinement (GAIR) to extract superquadrics (bottom row).}
\label{fig:parade}
\end{figure*}

The task of \emph{fitting geometric models} to visual data to derive structured interpretations of unorganized spatial points is a key challenge in Computer Vision and Graphics. 
In the context of 3D data, \emph{primitive decomposition} becomes particularly relevant, as geometric model fitting enables representing complex scenes using simple primitives such as planes, cylinders, or superquadrics, providing a compact and interpretable approximation of real-world 3D structures as depicted in Figure \ref{fig:parade}.  
This  is fundamental for applications such as scene understanding and robotics, where accurate shape modelling underpins tasks such as object grasping~\cite{wu2025autonomous}, collision avoidance~\cite{dhal_collision}, and navigation~\cite{Zhang2023ALA}. 
More generally, primitive-based representations are attractive because they offer compact, interpretable descriptions of 3D shape, often requiring only a small number of parameters to represent structures that would otherwise require dense representations such as voxel grids, point clouds or meshes.

However, to be effective, primitive decomposition must balance compactness with geometric accuracy. A useful decomposition should represent the scene with a small number of primitives while still capturing its relevant geometric structure. This is challenging: not only must the method be robust to noise and outliers, which are almost unavoidable in real data, but the simultaneous presence of multiple primitives requires jointly estimating both segmentation and model parameters, significantly increasing the complexity of the problem. This challenge is further exacerbated in scenes where primitives are spatially adjacent and partially overlapping.

One of the most widely adopted approaches for robust model fitting is \emph{consensus maximization}, as epitomized by the well-known \ransac~\cite{RANSAC} paradigm and its numerous variants. \ransac estimates a model by iteratively sampling subsets of the data and selecting the hypothesis that maximizes the number of inliers according to a residual threshold. This strategy has proven highly effective for single-model fitting, particularly in the presence of noise and outliers. 

Several extensions have been proposed to improve the quality of the estimated models. Methods such as \loransac~\cite{LO} and \gcransac~\cite{GCRANSAC} incorporate local optimization and spatial coherence, refining inlier sets by encouraging neighboring points with low residuals to be jointly selected as inliers, while penalizing assignments that are inconsistent with the local data structure. These approaches improve robustness and convergence, and represent the current standard in many geometric estimation tasks.

In parallel, multi-model fitting methods extend this paradigm to estimate multiple structures. Approaches such as \ransacov~\cite{magri2016multiple} formulate the problem as a model selection or coverage optimization task, where a pool of candidate models is generated, and a subset is selected to maximize the number of explained points.

Despite their effectiveness, these approaches are primarily designed for settings where model fitting can be reliably driven by residual-based criteria and spatial proximity. 
However, this assumption becomes limiting in 3D primitive decomposition, where the underlying structures are continuous surfaces characterized by local geometric properties such as orientation and smoothness.

In this setting, spatial proximity becomes a weak proxy for surface membership: points that are close in Euclidean space may belong to different primitives, especially near regions where surfaces intersect or come into contact. As a result, formulations based solely on residuals and proximity may incorrectly propagate inliers across adjacent structures.

\begin{figure*}
\centering

 \begin{subfigure}[t]{0.245\textwidth}
        \centering
        \includegraphics[width=\linewidth,trim={0cm 0cm 0cm 0cm},clip]{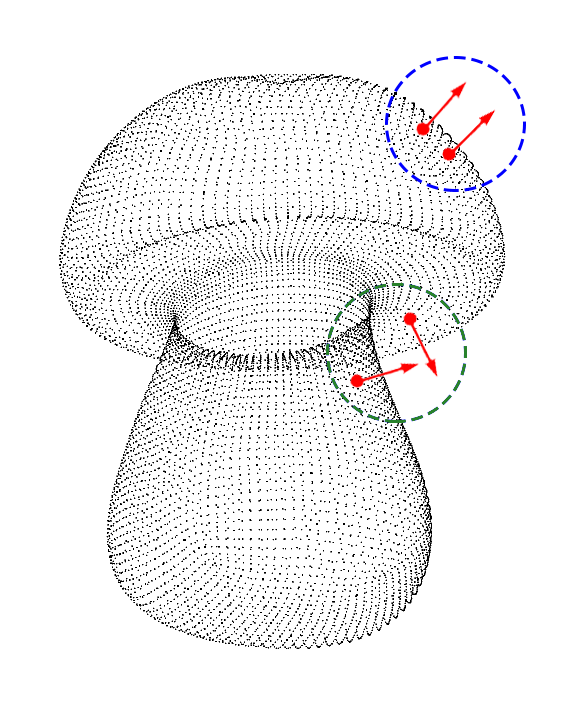}
        \caption{Geometric ambiguity}
        \label{fig:idea_a}
    \end{subfigure}
 \begin{subfigure}[t]{0.245\textwidth}
        \centering
        \includegraphics[width=\linewidth]{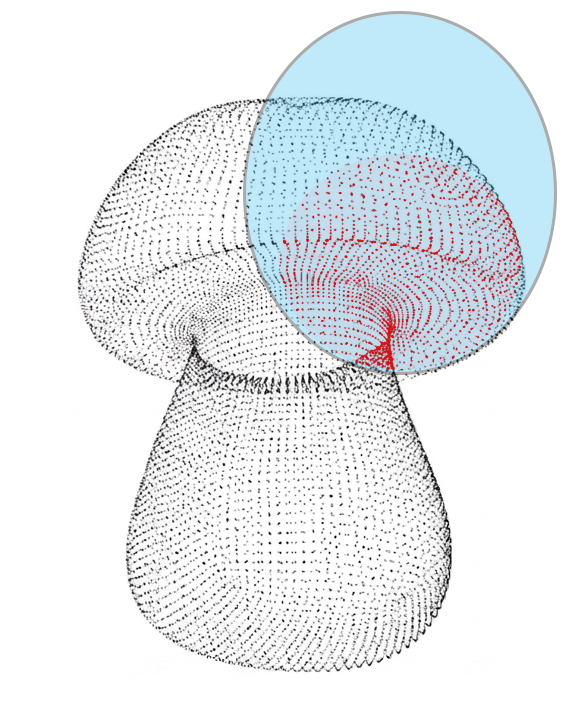}
        \caption{Residual based}
        \label{fig:idea_b}
\end{subfigure}
 \begin{subfigure}[t]{0.245\textwidth}
        \centering
        \includegraphics[width=\linewidth]{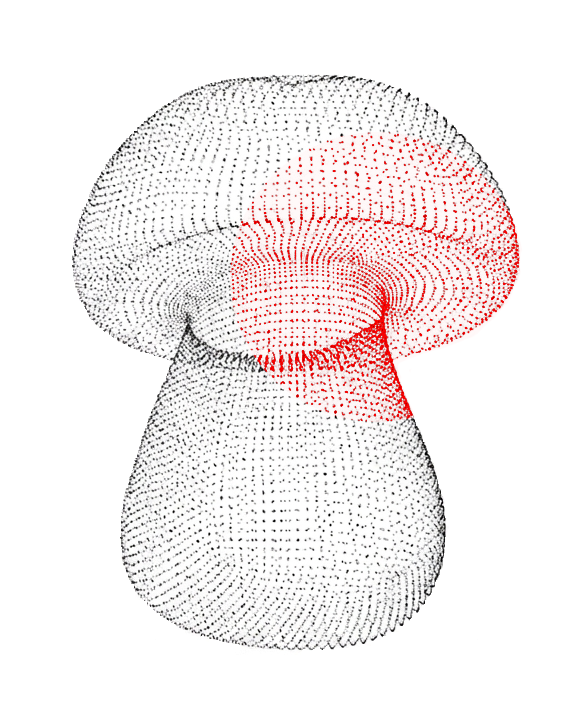}
        \caption{Proximity based}
        \label{fig:idea_b}
\end{subfigure}
 \begin{subfigure}[t]{0.245\textwidth}
        \centering
        \includegraphics[width=\linewidth]{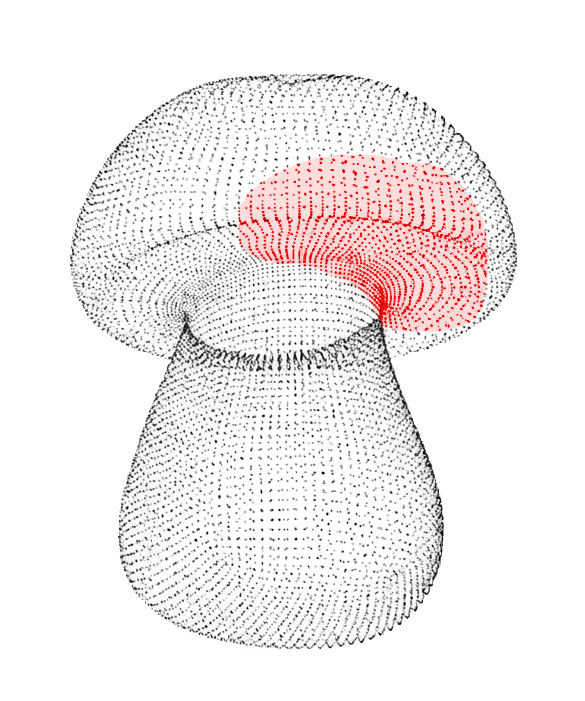}
        \caption{Geometric-aware}
        \label{fig:idea_b}
\end{subfigure}

\caption{Geometric ambiguity in 3D primitive decomposition and inlier refinement strategies. 
(a) Two spatially close points may belong to different surface regions, as indicated by their differing normals in the green circle. 
(b) Initial inlier set $\mathcal{I}_j$ obtained via residual-based thresholding. 
(c) Proximity-based refinement propagates inliers across adjacent surfaces due to proximity. 
(d) Incorporating geometric consistency prevents assignments across incompatible regions, yielding a coherent inlier set.}
\label{fig:idea}
\end{figure*}

This is illustrated in Figure~\ref{fig:idea}(a), where two pairs of points are shown at comparable spatial distance. 
In the first pair (highlighted by the blue dashed circle), the points exhibit consistent surface orientations and can plausibly belong to the same primitive. 
In contrast, the second pair (highlighted by the green circle) has different normals, indicating that the points belong to different surface regions, despite being spatially close. 
This example highlights that spatial proximity alone is insufficient to determine surface membership. Instead, proximity and local geometry must be considered jointly: while similar normals do not guarantee that two points lie on the same primitive, nearby points with inconsistent normals should not be assigned to the same structure.

To illustrate this effect, let $h_j$ denote a candidate model hypothesis generated by a consensus maximization procedure, and $\mathcal{I}_j$ the corresponding inlier set obtained via residual-based thresholding. 
Figure~\ref{fig:idea}(b–d) shows how different refinement strategies affect the resulting inlier set: spatial coherence alone may incorrectly expand $\mathcal{I}_j$ across geometrically distinct regions, whereas incorporating geometric consistency prevents such assignments and yields a more coherent set of inliers.

To address this limitation, we propose a \emph{Geometric-Aware Inlier Refinement}  strategy (GAIR) that explicitly accounts for the local surface structure of the data. 
Our key idea is to augment standard residual-based formulations with geometric priors that capture surface-level properties, such as normal consistency and local surface coherence. By incorporating these cues, inlier selection is no longer driven solely by spatial proximity, but also by the coherence of points with respect to the underlying surface.

We formulate this refinement as an energy minimization problem over a labeling of the point cloud, which is efficiently solved via graph-cut optimization. 
Building on existing energy-based formulations (e.g., GC-\ransac), we introduce geometric priors tailored to 3D primitive decomposition, enabling inlier selection consistent with the underlying surface structure. 
The resulting framework can be integrated into standard consensus maximization pipelines, improving the stability of the estimated models and reducing erroneous inlier propagation across adjacent primitives.

In this work, we focus on \emph{superquadric primitives}, which provide a compact and expressive representation capable of modeling a wide range of shapes with a small number of parameters. Their flexibility makes them particularly appealing for primitive-based scene representations.

However, this expressiveness comes at the cost of increased estimation complexity. Unlike simpler primitives, superquadrics cannot be reliably inferred from minimal samples in closed form, as their parameters must be recovered through a non-linear optimization process. As a result, their estimation is highly sensitive to noise, outliers, and initialization, making robust fitting particularly challenging in realistic settings. This makes superquadric fitting a  suitable testbed for evaluating inlier-refinement strategies in 3D.

\subsection{Contributions.} 

This work presents a geometric-aware formulation for robust primitive fitting in 3D point clouds. Specifically, we observe that, while consensus maximization methods have been successfully extended with spatial coherence (\emph{e.g.}, GC-\ransac), their formulation remains primarily driven by point-wise residuals and proximity, which do not explicitly capture the underlying surface geometry in 3D primitive decomposition. In this respect, our contributions are:
\begin{itemize}
    \item \textbf{Geometric-Aware Inlier Refinement (GAIR).}
    We propose a novel inlier refinement strategy formulated as an energy minimization problem, extending standard graph-cut formulations by incorporating geometric priors with a particular focus on normal consistency and surface-aware regularization. This enables inlier selection that is coherent with the underlying surface geometry, going beyond purely residual- and proximity-based criteria.

    \item \textbf{GAIR-\ransac for single-model fitting.}
    We integrate the proposed refinement into a consensus maximization framework, resulting in a robust estimator for superquadric fitting. The geometric-aware refinement improves the stability of the inlier set, leading to more accurate parameter estimation and a favorable accuracy–runtime trade-off compared to existing \ransac-based methods.

    \item \textbf{GAIR-\ransacov for primitive decomposition.}
    We extend the framework to the multi-model setting by combining GAIR-based hypothesis generation with a Maximum Coverage formulation for model selection. This solution enables robust decomposition of point clouds into multiple superquadrics, particularly in challenging scenarios with adjacent or overlapping primitives.
\end{itemize}

This work extends our previous conference version~\cite{ferraris2025geometric} by introducing the extension to multi-model primitive decomposition and a more comprehensive evaluation.
We validate the proposed approach on synthetic, controlled scenarios \textcolor{black}{and real data} demonstrating improved robustness and stability compared to existing methods. While the present study focuses on relatively structured settings, the formulation provides a foundation for tackling more complex real-world primitive decomposition tasks.

\section{Problem formulation}
\label{sec:probForm}

 We consider the problem of \emph{primitive decomposition} in 3D point clouds. 
A scene is represented by an unordered set of noisy points 
$\mathcal{D} = \{\vec{x}_1, \dots, \vec{x}_N\}$, with $\vec{x}_i \in \mathbb{R}^3$. 
For each point $\vec{x}_i$, a normal vector $\vec{n}_i$ is either provided or estimated, yielding a set 
$\textcolor{black}{\mathcal{V}} = \{\vec{n}_1, \dots, \vec{n}_N\}$, where each $\vec{n}_i \in \mathbb{R}^3$ is associated with a point in $\mathcal{D}$ \textcolor{black}{and is normalized so that $\lVert\mathbf{n}_i\rVert_2=1$. We assume that all the normals are consistently oriented.}
The goal is to recover a set of geometric primitives that explain the observed data, represented as a collection of \textcolor{black}{parameters $\{\theta_1, \dots, \theta_\kappa\}$, each defining a superquadric,} where the number of primitives $\kappa$ may be either specified or estimated as part of the decomposition process.

The inlier threshold $\varepsilon$, given a primitive, defines its inliers as those points with residuals lower than $\varepsilon$.
\textcolor{black}{As customary in consensus-based robust estimation, we assume that  $\varepsilon$ is provided as input as an estimate of the expected measurement uncertainty.} 
While we focus on superquadrics, the formulation can be extended to other primitives (e.g., planes, spheres or collections of surface patches).

\paragraph{Superquadrics.}

Superquadrics, introduced in~\cite{Barr}, are a family of parametric surfaces derived from quadrics that can represent a wide range of shapes using a compact set of parameters. 
Their flexibility arises from the use of exponent parameters that control the roundness or sharpness of the surface along different directions.

A canonical superquadric is defined in its local coordinate system by the implicit function:
\begin{equation}
\label{eq:superquadricImplicit}
F(\mathbf{x}, \Lambda) =
\left(
\textcolor{black}{\left|\frac{x}{a_1}\right|}^{\frac{2}{\varepsilon_2}}
+
\textcolor{black}{\left|\frac{y}{a_2}\right|}^{\frac{2}{\varepsilon_2}}
\right)^{\frac{\varepsilon_2}{\varepsilon_1}}
+
\textcolor{black}{\left|\frac{z}{a_3}\right|}^{\frac{2}{\varepsilon_1}}
= 1 \,,
\end{equation}
where $\mathbf{x} = (x,y,z)^\top \in \mathbb{R}^3$.

The parameter set $\Lambda = \{a_1, a_2, a_3, \varepsilon_1, \varepsilon_2\}$
defines the canonical shape, where $a_1, a_2, a_3$ control the scale along
the coordinate axes, and $\varepsilon_1, \varepsilon_2$ control the shape.
In particular, $\varepsilon_1$ governs the geometry along the $z$-axis,
while $\varepsilon_2$ affects the cross-section in the orthogonal plane.

To represent a superquadric in the scene, the canonical model is mapped
from its local coordinate system to the world coordinate system via a rigid
transformation. This transformation is parameterized by a rotation
$R(\alpha,\beta,\gamma)$ and a translation
$\mathbf{t} \in \mathbb{R}^3$. The full model is therefore defined by
$\theta = \{\Lambda, \alpha, \beta, \gamma, \mathbf{t}\}$, resulting in a
total of 11 parameters (5 shape parameters and 6 pose parameters).

Given a superquadric $\theta$, measuring the residual $r(\mathbf{x},\theta)$ between a point $\mathbf{x} \in \mathbb{R}^3$ and the surface is non-trivial, and different approximations can be used. 
Since the implicit function $F(\cdot,\Lambda)$ is defined in the canonical coordinate system, a point $\vec{x}$ in the scene is first mapped to canonical coordinates as
$
\mathbf{x}' = R^\top(\vec{x} - \mathbf{t})$,
where $R$ and $\mathbf{t}$ are the rotation and translation associated with $\theta$.

We use $r$ as residual distance: the  \emph{radial distance} which projects $\vec{x}'$ onto the surface along the ray from the primitive center, measuring how much the point must be scaled to satisfy $F(\vec{x}',\Lambda)=1$. 
\textcolor{black}{
Since $F(t\mathbf{x}',\Lambda) = t^{2/\varepsilon_1}F(\mathbf{x}',\Lambda)$ for $t>0$, the point on the surface lying along the ray through $\mathbf{x}'$ is $t^\star \mathbf{x}'$ with $t^\star = F(\mathbf{x}',\Lambda)^{-\varepsilon_1/2}$. The \emph{radial distance} residual is then given in closed form by the positive quantity
\begin{equation}
\label{eq:radial_residual}
r(\mathbf{x},\theta) = \lVert \mathbf{x}'\rVert\,\Big|\,1 - F(\mathbf{x}',\Lambda)^{-\varepsilon_1/2}\Big|,
\end{equation}
where $\mathbf x' = R^\top(\mathbf x - \mathbf t)$. The signed quantity $\lVert\mathbf x'\rVert\big(1-F(\mathbf x',\Lambda)^{-\varepsilon_1/2}\big)$ is positive outside the surface and negative inside it.}

This formulation is geometrically intuitive and accurate for convex shapes, but becomes less reliable for highly non-convex configurations ($\varepsilon_1$ or $\varepsilon_2$ >> 1), where the projection may deviate from the true closest point.

\section{Related Work}
\label{sec:relWork}

We review two closely related research directions. On one side, our work addresses the general problem of robust model estimation, which is traditionally studied within the \ransac paradigm and its numerous extensions. On the other hand, it tackles the task of geometric primitive fitting, aiming to approximate complex shapes exploiting parametric structures such as planes, cylinders, or superquadrics.
\paragraph{Robust model fitting.}
Robust model estimation is commonly formulated as a consensus maximization problem, where the goal is to identify a model hypothesis $h_j$ that maximizes the size of its inlier set
\begin{equation}
\label{eq:residual_based_inliers}
\mathcal{I}_j = \{x_i \in \mathcal{D} \mid r(x_i, h_j) < \varepsilon\},
\end{equation}
defined in terms of a residual function $r(\cdot)$ and inlier threshold $\varepsilon$.

The \ransac paradigm~\cite{RANSAC}, originally introduced in the context of model fitting for cartographic data, addresses this problem by iteratively sampling minimal subsets of the data and selecting the hypothesis that yields the largest consensus. Since its introduction, \ransac has become a cornerstone of robust estimation, particularly in computer vision tasks such as image matching, visual localization, and 3D reconstruction.

Several extensions have been proposed to improve the quality of the estimated models. LO-\ransac~\cite{LO} incorporates a local optimization step that iteratively refines the model using the current inlier set. GC-\ransac~\cite{GCRANSAC} further extends this approach by introducing spatial regularization through a graph-cut formulation. In this setting, model fitting can be interpreted as a \emph{labelling problem}, where each point is assigned to either the inlier or outlier class.

More formally, GC-\ransac defines an energy function composed of a \emph{data fidelity term} $E_1$, based on the residual $r(\mathbf{x}_i,h_j)$, and a \emph{smoothness term} $E_2$, which promotes spatial coherence by encouraging neighboring points with low residuals to be jointly selected as inliers.

As a result, the initial consensus set $\mathcal{I}_j$ is refined into a spatially coherent set, as illustrated in Figure~\ref{fig:idea}(c). This proximity propagation regularizes the estimation and is particularly effective in settings where proximity is a reliable proxy for model membership, such as feature matching and two-view geometry.
However, as discussed in Section~\ref{sec:introduction}, this assumption becomes less informative in 3D primitive decomposition, where spatial proximity does not necessarily reflect the underlying surface structure.

\paragraph{Multi-model fitting.}
Multi-model fitting extends consensus maximization to the simultaneous estimation of multiple structures, requiring both model selection and data segmentation. 
Formally, this corresponds to jointly estimating a set of models $\{\theta_k\}_{k=1}^{\kappa}$ and assigning each point $x_i \in \mathcal{D}$ to one of them, or to an outlier class.

Clustering-based approaches, such as T-Linkage~\cite{Mag14}, group points according to shared model preferences, even considering multiple classes of parametric models~\cite{magri2019fitting, magri2021multilink}.
These methods naturally produce a partition of the data $\mathcal{D}$, but are typically greedy and often require additional post-processing to handle the presence of overlapping structures.
\textcolor{black}{A related family of methods performs \emph{hierarchical} segmentation, progressively merging or splitting surface regions to build a multi-resolution decomposition rather than committing to a single partition. Attene \emph{et al.}~\cite{attene2010hierarchical} recover a hierarchical structure of point-sampled surfaces, while Zhang \emph{et al.}~\cite{zhang2015hierarchical} perform hierarchical mesh segmentation guided by quadric surface fitting. These methods  do not require the number of segments to be specified in advance. Because their merging or splitting decisions are typically greedy, early local decisions may produce suboptimal boundaries.
}
Optimization-based methods instead formulate multi-model fitting as a global energy minimization problem. 
Approaches such as \textsc{Pearl}~\cite{PEaRL}, \textsc{Prog-X}~\cite{progx}, and \textsc{Multi-X}~\cite{multix} jointly estimate model assignments by balancing a data fidelity term, typically expressed through residuals $r(\vec{x}_i,h)$, with regularization terms enforcing spatial coherence. 
In particular, \textsc{Pearl} constructs a graph over the data points and optimizes the assignment via graph-cut techniques (e.g., $alpha$-expansion), similarly to other energy-based formulations.
These approaches encourage neighboring points to share the same assignment, effectively promoting spatial consistency in the inferred inlier sets $\mathcal{I}_j$. However, the resulting assignments are typically hard and remain primarily driven by residuals and proximity, which can lead to incorrect labeling in regions where multiple primitives intersect or lie in close spatial proximity.

\ransacov~\cite{magri2016multiple} takes a different perspective by formulating model selection as a maximum coverage problem, selecting a subset of candidate models that maximizes the number of explained points. 
While this formulation can better handle overlapping structures, it still relies on residual-based inlier definitions.

Despite their differences, all these approaches fundamentally rely on residual-based criteria and spatial coherence, which may lead to incorrect assignments in the presence of adjacent or intersecting primitives, where proximity does not reflect the underlying surface structure.

\paragraph{Primitive fitting.}

Primitive fitting can be seen as a specific instance of model fitting in which the goal is to approximate complex shapes using parametric primitives such as planes, cylinders, or superquadrics. 
 Among the possible primitives, superquadrics have attracted significant attention for their ability to represent a wide range of shapes with a relatively small number of parameters.

Early works introduced superquadrics for modeling objects in range images using least-squares fitting~\cite{Solina}, and explored different error measures $r(\vec{x},\theta)$ for single superquadric recovery~\cite{Gross}. Notably, Schnabel \emph{et al.}~\cite{schnabel2007efficient} proposed an efficient \ransac-based approach for primitive detection that exploits surface normals, using them for inlier validation through both distance and angular thresholds. 
In particular, a point $\vec{x}_i$ is considered an inlier \textcolor{black}{of a superquadric $h_j$} if it satisfies both a residual constraint and an angular consistency constraint:
{\color{black}
\begin{equation*}
\label{eq:normal_inlier}
\mathcal{I}_j =
\left\{
\mathbf{x}_i \in \mathcal{D}
\;\middle|\;
r(\mathbf{x}_i,h_j) < \varepsilon,
\arccos\!\left(
    \left|
        \mathbf{n}_i^{\top}
        \mathbf{n}_{h_j}(\mathbf{x}_i)
    \right|
\right)
< \alpha
\right\}
\end{equation*}
}

where $\mathbf{n}_{h_j}(\mathbf{x}_i)$ denotes the unit normal of model
$h_j$, evaluated at the projection of $\mathbf{x}_i$ onto the model
surface, and $\alpha$ is the angular threshold. In contrast, our
formulation incorporates normal information directly into the
optimization through both unary penalties and pairwise consistency
terms. Consequently, geometric information actively influences inlier
selection rather than being used solely as a post-fitting validation
criterion.

Subsequent approaches explored alternative estimation strategies. Solina \emph{et al.}~\cite{SuperSelectRecovery} proposed a recover-and-select paradigm, where models are iteratively refined by progressively adding nearby points and then evaluated through an objective function. Liu \emph{et al.}~\cite{Probsuper} introduced a probabilistic framework based on Expectation Maximization (EM), explicitly modeling noise and outliers to improve robustness.

More recent methods have investigated different directions. Monnier \emph{et al.}~\cite{Nerfsuperquadric} proposed Differentiable Blocks World, which optimizes textured superquadric primitives from multi-view images via differentiable rendering, achieving high reconstruction fidelity at the cost of significant computational complexity. Ramamonjisoa \emph{et al.}~\cite{MonteBoxFinder} introduced a stochastic search method for fitting primitives to noisy point clouds, handling missing data and noise without requiring training data, though it is limited by the use of predefined primitive types such as cuboids.

Learning-based approaches have also gained momentum, proposing data-driven methods that infer collections of 3D primitives (e.g., superquadrics~\cite{Desuperq, Pas20} or cuboids~\cite{Tulsiani}) from clean training data. While promising, these approaches typically require large annotated datasets and may struggle to generalize across different domains or primitive families. In contrast, our work follows a purely algorithmic direction and builds upon our previous formulation \cite{ferraris2025geometric}, extending it to the multi-model fitting setting and enabling robust primitive decomposition in complex scenes.

\section{Method}

Our approach is built around a geometric-aware inlier refinement strategy (GAIR), which serves as the core component of the proposed framework. We develop this idea at three complementary levels.

First, in Section \ref{sec:gair} we introduce GAIR as a standalone inlier refinement module, formulated as an energy minimization problem that integrates geometric priors such as normal consistency and \textcolor{black}{local surface coherence}. This component improves the quality of inlier sets associated with individual model hypotheses.

Second, we embed GAIR within a consensus maximization pipeline, resulting in GAIR-\ransac, a robust estimator for single-model superquadric fitting as detailed in Section \ref{sec:gair_ransac}. This integration improves both the stability of model estimation and the convergence of the RANSAC procedure.

Finally, in Section \ref{sec:gair_ransacov}, we extend the framework to the multi-model setting through GAIR-\ransacov. In this case, GAIR generates a pool of candidate hypotheses, which are subsequently selected via a global Maximum Coverage formulation to obtain a consistent primitive decomposition. The following sections describe these components in detail.

\subsection{Geometric-Aware Inlier Refinement}
\label{sec:gair}
\begin{figure*}[!ht]
\centering
\includegraphics[width=\linewidth]{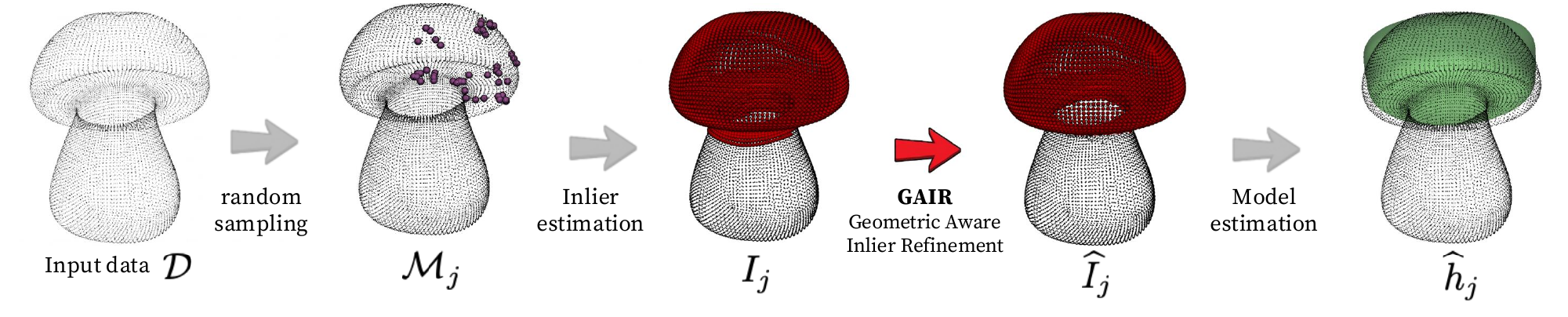}
\caption{Our geometric-aware refinement in a nutshell. Given a candidate model $h_j$ and its initial inlier set $\mathcal{I}_j$, GAIR produces a refined set $\widehat{I}_j$ that better adheres to the underlying surface geometry.}
\label{fig:hl-pipeline}
\end{figure*}

Consensus maximization approaches generate a pool of candidate models $\{h_j\}$ by sampling subsets $\mathcal{M}_j \subset \mathcal{D}$ and fitting a geometric model to each sample. The final solution consists of selecting the subset $\mathcal{S}$ of hypotheses that best explain the data in terms of their associated inlier sets (see Section~\ref{sec:relWork}).
As illustrated in Figure~\ref{fig:hl-pipeline}, given a sampled subset $\mathcal{M}_j$, a model $h_j$ is estimated and an initial set of inliers $\mathcal{I}_j$ is obtained as in \eqref{eq:residual_based_inliers} through residual thresholding. While this step effectively identifies points that are close to the model, it does not explicitly account for the underlying surface structure.

The goal of our \emph{Geometric-Aware Inlier Refinement} (GAIR) is to refine this inlier set. Starting from a candidate model $h_j$ and its initial inliers $\mathcal{I}_j$, we compute a refined set $\widehat{I}_j$ that better adheres to the true geometric structure of the data. In particular, the refinement step removes geometrically inconsistent points and enforces local coherence. As shown in the example, the refined model better adheres to the cap of the mushroom. This leads to more accurate model estimates and \textcolor{black}{provides a favorable accuracy--runtime trade-off within the fitting pipeline}, as reported in our experiments (see Figure~\ref{fig:single_model_me_time}).

The key observation is that inlier sets should be coherent at the \emph{surface level}: points belonging to the same primitive are expected not only to have low residuals, but also to exhibit consistent local geometry, such as aligned normals.
To formalize this idea, we reinterpret the inlier set associated with the $j$-th model as a binary labeling over the point cloud:
\begin{equation}
\label{eq:initial_labeling}
f_j : \mathcal{D} \rightarrow \{0,1\},
\end{equation}
where $f_j(\vec{x}) = 1$ if $\vec{x}$ is classified as an inlier, and $0$ otherwise.

Given the initial labeling induced by $\mathcal{I}_j$, our objective is to compute a refined labeling $\hat{f}_j$, defining an updated inlier set:
\begin{equation}
\label{eq:inlier_label}
\hat{\mathcal{I}}_j = \{ \vec{x} \in \mathcal{D} \mid \hat{f}_j(\vec{x}) = 1 \}.
\end{equation}

Building upon the energy minimization framework of GC-\ransac, we cast this refinement as a labeling problem over a graph defined on the point cloud $\mathcal{D}$, where both data fidelity and geometric consistency are taken into account.

\subsubsection{Energy Formulation}

For clarity, we omit the model index and consider a generic labeling function $f$. We denote by $f_p \in \{0,1\}$ the label assigned to a point $\vec{p}$, i.e., $f_p = f(\vec{p})$.

We define the energy associated with a labeling $f$ over the point cloud graph as:
\begin{equation}
\label{eq:energy}
E(f) = \sum_{\vec{p} \in \mathcal{D}} E_1(f_p) + \sum_{(\vec{p}, \vec{q}) \in \mathcal{E}} E_2(f_p, f_q),
\end{equation}
where $\mathcal{E}$ denotes the set of edges connecting neighboring points (details on how to construct the graph are provided in Sec.\ref{sec:graph}).

The energy consists of two terms:
\begin{itemize}[noitemsep]
    \item $E_1$ is a unary term encoding data fidelity, measuring how well each point agrees with the model {\color{black}in terms of both geometric residual and normal orientation}. This corresponds to the standard residual-based criterion used in \ransac.
    \item $E_2$ is a pairwise term promoting coherence between neighboring points.
\end{itemize}

\paragraph{Unary term (data fidelity).}
The unary term measures how well each point fits the model:
\begin{equation}
\label{eq:unary}
E_1(f_p) =\begin{cases}\bar{d}_p
{\color{black}\displaystyle+\frac{1}{2}(1-{\vec{n}}_p \cdot{\vec{n}}_{h_j}(\vec{p}))
}
& \text{if } f_p = 1, \\
1, & \text{if } f_p = 0.
\end{cases}
\end{equation}
{\color{black}Here, $\hat{\vec{n}}_p$ denotes the normalized observed unit normal associated with point $\vec{p}$, while $\hat{\vec{n}}_{h_j}(\vec{p})$ denotes the unit surface normal predicted by model $h_j$ at $\vec{p}$}
and where
\begin{equation}
\bar{d}_p =
\min\left(\frac{{\color{black}|r(\vec{p},h_j)|}}{\varepsilon},1\right).
\end{equation}

is the  residual of an inlier from model $h_j$ normalized by the inlier threshold $\varepsilon$ and  truncated to $1$. This normalization ensures that all residuals lie in $[0,1]$, which is crucial to balance the unary and pairwise contributions in the energy. In addition, truncation limits the influence of outliers and guarantees bounded costs, a desirable property for stable graph-cut optimization.

\paragraph{Pairwise term (geometric consistency).}
The pairwise term encourages neighboring points to share the same label when they are geometrically compatible.
We first define a \emph{normal-based coherency measure} between neighboring points $p$ and $q$:
\begin{equation}
\label{eq:normal_coherence}
C(\vec{n}_p, \vec{n}_q) = \frac{1}{2}(1 + \vec{n}_p \cdot \vec{n}_q),
\end{equation}
which lies in $[0,1]$ and captures the alignment between surface normals. 

A basic formulation penalizes label disagreement proportionally to this coherency:
\begin{equation}
E_2(f_p, f_q) =
\begin{cases}
C(\vec{n}_p, \vec{n}_q), & \text{if } f_p \neq f_q, \\
0, & \text{if } f_p = f_q.
\end{cases}
\end{equation}
This encourages neighboring points with aligned normals to share the same label, promoting surface-consistent inlier sets. 
To improve robustness, we further modulate the pairwise term using the point-to-model residuals. Let $\bar{d}_p$ and $\bar{d}_q$ denote the normalized distances of points $p$ and $q$ from the model. The final pairwise term is defined as:
\begin{equation}
\label{eq:pairwise}
E_2(f_p, f_q) =
\begin{cases}
C(\vec{n}_p, \vec{n}_q), & \text{if } f_p \neq f_q, \\
0, & \text{if } f_p = f_q = 1, \\
\left(1 - \frac{\bar{d}_p + \bar{d}_q}{2}\right)C(\vec{n}_p, \vec{n}_q), & \text{if } f_p = f_q = 0,
\end{cases}
\end{equation}
{\color{black}
where $\bar{d}_p$ is defined as:  
\begin{equation}
\bar{d}_p =
\min\left(\frac{|r(p, h_j)|}{\lambda\varepsilon}, 1\right),
\end{equation}
where $\lambda$ is an outlier scale factor. The term $\bar{d}_q$ is defined accordingly.  In our implementation we set
$\lambda=3$.
}

This formulation of the pairwise term introduces two complementary effects: on the one hand neighboring points with aligned normals are encouraged to share the same label. On the other hand, points far from the model have a reduced influence on the labeling, preventing outliers from enforcing incorrect smoothness.
The distances $\bar{d}_p$ and $\bar{d}_q$ are normalized and bounded to ensure that the pairwise term remains within $[0,1]$.

Moreover, the pairwise term \eqref{eq:pairwise} satisfies the submodularity condition required by graph-cut optimization:
\begin{equation}
    \left(
        1-\frac{\bar{d}_p+\bar{d}_q}{2}
    \right)
    C(\mathbf{n}_p,\mathbf{n}_q)
    \leq
    2C(\mathbf{n}_p,\mathbf{n}_q),
\end{equation}
since the residual term is bounded in $[0,1]$.

{\color{black}
In practice, we construct the graph by retaining only edges whose normal coherence satisfies
\begin{equation}
    C(\mathbf{n}_p,\mathbf{n}_q)>0.9.
\end{equation}
Edges with $C(\mathbf{n}_p,\mathbf{n}_q)\leq 0.9$ are discarded by setting their weights to zero. This choice
 retains connections only between neighboring points whose normals differ by less than $37^\circ$. This conservative graph construction restricts the smoothness prior to locally coherent surface regions.

\paragraph{Remark.}
The pairwise term $E_2$ is the only component of the proposed energy that couples the labels of neighboring points. Indeed, if $E_2 \equiv 0$, Eq.~\eqref{eq:energy} reduces to the purely unary formulation
\begin{equation}
E(f)=\sum_{\vec{p}\in\mathcal D}E_1(f_p),
\end{equation}
which can be minimized independently for each point. The resulting labeling, however, does not generally coincide with the consensus set of vanilla \ransac, since the proposed unary term evaluates both the point-to-model residual and the agreement between the observed and model normals. Standard residual-based \ransac is recovered only as the further special case in which the normal-consistency contribution is removed from $E_1$.
}

If the pairwise term is defined using only residual-based quantities, without normal-based geometric consistency, it reduces to the proximity based formulation used to refine the inlier sets in GC-\ransac. 
In this case, the pairwise term can be written as:
\begin{equation}
\label{eq:gc_pairwise}
E_2^{\text{GC}}(f_p, f_q) =
\begin{cases}
1, & \text{if } f_p \neq f_q, \\
\frac{1}{2}(\bar{d}_p + \bar{d}_q), & \text{if } f_p = f_q = 1, \\
1 - \frac{1}{2}(\bar{d}_p + \bar{d}_q), & \text{if } f_p = f_q = 0,
\end{cases}
\end{equation}

Our GAIR refinement extends this framework by incorporating normal-based geometric consistency and residual-aware weighting within the pairwise term, enabling surface-aware inlier refinement.

\subsubsection{Graph Construction and Optimization}
\label{sec:graph}
To minimize the energy in Eq.~\eqref{eq:energy}, we formulate the problem as a binary labeling task on a graph $\mathcal{G}$ and solve it using the graph-cut algorithm.

\paragraph{Graph construction.}
We construct an undirected graph $\mathcal{G} = (\mathcal{D}, \mathcal{E})$ from the input point cloud. Each node $p \in \mathcal{D}$ corresponds to a point in $\mathcal{D}$, and edges $(p,q) \in \mathcal{E}$ connect neighboring points.
The neighborhood structure is defined locally: for each point $p$, we identify a set of neighboring points using a spatial criterion (e.g., $k$-nearest neighbors). This results in a sparse graph capturing the local geometric structure of the data.
\textcolor{black}{
The parameter $k$ controls the effective spatial support of the
regularization: for a fixed value of $k$, a higher sampling density
results in a smaller neighborhood radius. Increasing $k$ therefore
helps preserve stable local connectivity and reduces sensitivity to
non-uniform sampling, while keeping the neighborhood sufficiently
local to avoid connections across distinct surfaces.
In our implementation, the graph is built using a KD-tree-based
$k$-nearest-neighbor search. We use $k=6$ in the controlled
experiments, whereas for the more densely sampled real scans we use
a slightly denser graph with $k=10$.}
Each node is associated with unary costs derived from Eq.~\eqref{eq:unary}, while each edge is assigned a pairwise cost according to Eq.~\eqref{eq:pairwise}. Normal vectors $\vec{n}_p$ are either provided or estimated from the local neighborhood.
\textcolor{black}{For scanned point clouds, normals are estimated with Open3D using a $90$-nearest-neighbor local neighborhood. The estimated normals are then oriented through consistent tangent-plane propagation, and globally flipped when necessary to obtain an outward orientation.}

\paragraph{graph-cut optimization.}
The energy minimization problem is solved via a min-cut/max-flow procedure. The graph is augmented with two terminal nodes (source and sink), representing the inlier and outlier labels, respectively.

Pairwise terms define the weights of edges between neighboring nodes.

The minimum cut partitions the graph into two disjoint sets:
\begin{itemize}[noitemsep]
    \item nodes connected to the source correspond to inliers ($f(p) = 1$),
    \item nodes connected to the sink correspond to outliers ($f(p) = 0$).
\end{itemize}
This yields the optimal labeling $\hat{f}$ that minimizes the energy in Eq.~\eqref{eq:energy}.
The overall refinement procedure is summarized in Algorithm~\ref{alg:gair}.

\begin{algorithm}[t]
\caption{\textbf{GAIR}}
\textbf{Input:} point cloud $\mathcal{D}$, normals $\textcolor{black}{\mathcal{V}}$, candidate model $h_j$,\\
\hspace*{\algorithmicindent} initial inlier set $\mathcal{I}_j$,\\
\hspace*{\algorithmicindent} neighborhood graph $\mathcal{G} = (\mathcal{D}, \mathcal{E})$,\\
\textbf{Output:} refined inlier set $\widehat{\mathcal{I}}_j$
\label{alg:gair}
\begin{algorithmic}[1]
\State{Get the initial labeling $f_j$  from $\mathcal{I}_j$ using \eqref{eq:initial_labeling} }\label{alg:gair:init}
\State{Compute unary cost $E_1$ using \eqref{eq:unary}} \label{alg:gair:unary}
\State{Compute pairwise costs $E_2$ using \eqref{eq:pairwise}} \label{alg:gair:pairwise}
\State{$\widehat{f}_j =$ \textit{graphCut}($\mathcal{D}, \mathcal{G}, E_1, E_2$)} \label{alg:gair:cut}
\State{Compute $\widehat{\mathcal{I}}_j$ using \eqref{eq:inlier_label}} \label{alg:gair:extract}
\State{\textbf{return} $\widehat{\mathcal{I}}_j$}
\end{algorithmic}
\end{algorithm}
While the worst-case complexity of graph-cut is high, in practice it performs efficiently on sparse graphs arising from point cloud neighborhoods. The locality of the graph and the bounded energy terms lead to fast convergence, making the refinement step computationally tractable within a consensus maximization pipeline.

\subsection{GAIR-\ransac: Single-Model Superquadric Fitting}
\label{sec:gair_ransac}

We integrate the proposed \emph{Geometric-Aware Inlier Refinement} (GAIR) into a \ransac framework for the extraction of a single superquadric from a point cloud $\mathcal{D}$. The resulting procedure, summarized in Algorithm~\ref{alg:pipeline}, follows the classical hypothesize-and-verify paradigm, while introducing several adaptations to cope with the challenges of superquadric fitting.

At a high level, the algorithm iteratively samples, estimates, and refines a pool of $m$ candidate models, selecting the one $\theta^*$ that best explains the data in terms of its inlier support.

At each iteration $j$, a subset $\mathcal{M}_j \subset \mathcal{D}$ is sampled (line~\ref{algo:sample}) and used to estimate a model $h_j$ (line~\ref{algo:fit}). While in standard \ransac minimal samples are sufficient for closed-form models, superquadric estimation requires solving a non-linear optimization problem. As a result, minimal samples are often unstable and lead to poor initializations.
To address this issue, we adopt a \emph{local geometric sampling} strategy that favors  surface coherence. At each trial, a seed point is first sampled \textcolor{black}{randomly} from $\mathcal{D}$. Its normal vector is then used to define a local tangent plane, which induces a geometrically meaningful neighborhood. Specifically, we consider as candidate points those lying within the $k$-nearest neighbors of the seed and whose distance from the tangent plane is below a tolerance $\varepsilon$, thus restricting the pool to points that locally approximate the same surface patch. Finally, this set is further pruned by enforcing normal consistency, discarding points whose normals are not sufficiently aligned with that of the seed according to a cosine similarity threshold $\tau$.

From this pool, multiple candidate sample sets are generated and evaluated. Each candidate is constructed using Farthest-Point Sampling (FPS), initialized at the seed point, in order to ensure sufficient spatial coverage while remaining within a local surface patch. Since superquadric fitting is computationally expensive, we assess the quality of each candidate sample before model estimation. In particular, each sample is assigned a score that balances spatial compactness and geometric coherence. The sample achieving the best score is selected as $\mathcal{M}_j$.

Given the selected sample, the model $h_j$ is estimated via non-linear least squares, initialized using PCA to provide a stable starting point. Residuals are computed as the radial distance to the superquadric surface.

Once a candidate model is obtained, an initial inlier set $\mathcal{I}_j$ is computed via residual thresholding. If the current hypothesis improves over the best-so-far solution (line~\ref{algo:compare}), a local optimization step is triggered. Starting from $(h_j, \mathcal{I}_j)$, we iteratively refine the inlier set using GAIR (line~\ref{algo:GAIR_start}) and update the model parameters through an inner \ransac procedure.

The refinement loop continues until convergence, i.e., when no further improvement in the consensus is observed (lines~\ref{algo:GAIR_start}--\ref{algo:GAIR_end}). The final output is the model $h^*$ with the highest refined support.

\begin{algorithm}[h]
\caption{\textbf{GAIR-\ransac}}
\textbf{Input:} point cloud $\mathcal{D}$, threshold $\varepsilon$,\\
\hspace*{\algorithmicindent} maximum number of iterations $m$\\
\textbf{Output:} model parameters $\theta^*$
\label{alg:pipeline}
\begin{algorithmic}[1]

\State{$\mathcal{I}^* = \emptyset$, $h^* = \emptyset$}
\State{$\mathcal{G} = \textit{initGraph}(\mathcal{D})$, $\mathcal{V} = \textit{initNormals}(\mathcal{D})$}

\For{$j = 1 \rightarrow m$}

    \State \textbf{/* Local geometric sampling */}
    \State{$p_{seed} \sim \mathcal{D}$}
    \State{$\mathcal{P}_{local} = \textit{getNeighborhood}(p_{seed})$}
    \State{$\mathcal{P}_{filtered} = \textit{filterByNormals}(\mathcal{P}_{local}, \mathcal{V})$}

    \State \textbf{/* Candidate sample generation */}
    \State{$\{\mathcal{M}_j^k\} = \textit{generateCandidates}(\mathcal{P}_{filtered})$}
    
    \State \textbf{/* Sample scoring */}
    \For{each candidate $\mathcal{M}_j^k$}
        \State{$s_k = \textit{scoreSample}(\mathcal{M}_j^k)$}
    \EndFor
    \State{$\mathcal{M}_j = \arg\max_k s_k$} \label{algo:sample}

    \State \textbf{/* Model estimation */}
    \State{$h_j = \textit{fitSuperquadric}(\mathcal{M}_j, \text{PCA init})$} \label{algo:fit}

    \State{$\mathcal{I}_j = \textit{computeConsensus}(h_j, \varepsilon)$}

    \If{$|\mathcal{I}_j| > |\mathcal{I}^*|$} \label{algo:compare}

        \State{$terminate = \textbf{False}$}
        \State{$\mathcal{I} = \mathcal{I}_j$, $h = h_j$}

        \While{not $terminate$}

            \State{$\widehat{\mathcal{I}} = \textbf{GAIR}(\mathcal{D}, \mathcal{V}, h, \textcolor{black}{\mathcal{I}},\mathcal{G})$} \label{algo:GAIR_start}
            \State{$\widehat{h}, \widehat{c} = \textit{innerRANSAC}(\mathcal{D}, \widehat{\mathcal{I}}, \mathcal{I}, \varepsilon)$}

            \If{\textit{compareConsensus}($\mathcal{I}, \widehat{c}$)}
                \State{$\mathcal{I} = \widehat{c}$, $h = \widehat{h}$}
            \Else
                \State{$terminate = \textbf{True}$}
            \EndIf

        \EndWhile \label{algo:GAIR_end}

        \State{$\mathcal{I}^* = \mathcal{I}$, $h^* = h$}

    \EndIf

\EndFor

\State{\textbf{return} $h^*$}

\end{algorithmic}
\end{algorithm}

\subsection{GAIR-\ransacov: Primitive Decomposition}
\label{sec:gair_ransacov}

We now extend the proposed framework to the primitive decomposition, where the goal is to decompose a point cloud into \textcolor{black}{\emph{multiple superquadric hypotheses} $h_1,\ldots,h_\kappa$, parameterized respectively by $\theta_1,\ldots,\theta_\kappa$}. Unlike the single-model case, the multi-model problem requires disentangling multiple overlapping structures. We address this by decoupling hypothesis generation from model selection: first, we generate a pool of candidate models, and then we select the subset that best explains the data through a global optimization procedure that solves a Maximum Coverage problem~\cite{magri2016multiple}: select the set of $\kappa$ superquadrics that explain most of the points in $\mathcal{D}$.

\paragraph{Superquadric hypothesis generation strategy.}
Generating a large pool of superquadric hypotheses by purely random sampling would be computationally impractical. The superquadric fitting is expensive and highly sensitive to initialization, while the subsequent Maximum Coverage optimization becomes more demanding as the number of candidate models increases. For this reason, we adopt a structured strategy that produces a small set of high-quality hypotheses.

We use Sequential GAIR-\ransac as a structured hypothesis generator (Algorithm~\ref{alg:ransacov}), rather than as a final decomposition strategy.  At each run (lines~\ref{alg:run_start}--\ref{alg:run_end}), models are extracted sequentially: once a model $h$ is estimated using GAIR-\ransac (line~\ref{alg:gen_model}), its inliers are identified and removed from the current point set, and the procedure is repeated on the residual data until no sufficiently supported structure remains.

To improve the quality of the extracted hypotheses and remove spurious support, we adopt a more aggressive inlier removal strategy (line~\ref{alg:buffer}). After extracting a model, we remove not only its inliers, but also points in a neighborhood defined by a dilated threshold $\beta \cdot \varepsilon$, with $\beta > 1$. This suppresses residual support around already-explained structures, and can be interpreted as a mechanism to enforce diversity in the hypothesis pool, preventing multiple hypotheses from explaining the same local support.

Once a sequential run terminates, all points are restored (line~\ref{alg:reset}) and a new run is started from the original point cloud. To increase variability across runs, each run operates on a random subsampling of the input data (line~\ref{alg:subsample}). 

To control the size and quality of the hypothesis pool, we incorporate a coverage validation step (line~\ref{alg:cov}). After each extraction, a model is accepted only if its surface is sufficiently supported by the inlier set. In practice, we sample points uniformly on the estimated superquadric surface and compute the fraction of samples lying within distance $\varepsilon$ from the data. Models that fail to meet a minimum coverage threshold $\tau_{\text{cov}}$ are discarded (line~\ref{alg:skip}). This step reduces the number of redundant hypotheses, which is crucial since the complexity of the Maximum Coverage problem grows with the size of the hypothesis pool.

Repeating the sequential extraction process over multiple runs (lines~\ref{alg:run_start}--\ref{alg:run_end}) yields a hypothesis pool
$$
\mathcal{H} = \{h_1, \dots, h_M\},
$$
containing models that are generally well supported but still redundant, as the same underlying structure may be extracted multiple times with slightly different parameters.

\paragraph{Primitive selection via maximum coverage.}
Given the hypothesis pool $\mathcal{H}$, each model $h_j$ defines a geometric-aware consensus set $\widehat{\mathcal{I}}_j$ obtained via GAIR. The goal is to select a subset of models that jointly explains the largest number of points.
We cast this task as a Maximum Coverage problem~\cite{magri2016multiple}. Given a budget of $\kappa$ models, the objective is to select at most $\kappa$ hypotheses whose union of inlier sets maximizes the number of explained points:
\begin{equation}
\label{eq:maxcover}
\max_{\mathcal{S} \subseteq \mathcal{H},\ |\mathcal{S}| \leq \kappa} \left| \bigcup_{h_j \in \mathcal{S}} \widehat{\mathcal{I}}_j \right|.
\end{equation} 
\textcolor{black}{The selected set $\mathcal{S}$ provides} the desired primitive decomposition.

\begin{algorithm}[h]
\caption{\textbf{GAIR-\ransacov}}
\textbf{Input:} point cloud $\mathcal{D}$, threshold distance $\varepsilon$,\\
\hspace*{\algorithmicindent} number of models $\kappa$, number of runs $R$,\\
\hspace*{\algorithmicindent} coverage threshold $\tau_{\text{cov}}$, buffer factor $\beta$\\
\textbf{Output:} set of model parameters $\{\theta_k\}_{k=1}^{\kappa}$
\label{alg:ransacov}
\begin{algorithmic}[1]
\State{$\mathcal{H} = \emptyset$}
\For{$r = 1 \rightarrow R$} \label{alg:run_start}
    \State{$\mathcal{D}_{curr} = \textit{subsample}(\mathcal{D})$} \label{alg:subsample}
    \While{$\mathcal{D}_{curr}\neq \emptyset$}
        \State{$h, \widehat{\mathcal{I}} = \textbf{GAIR-\ransac}(\mathcal{D}_{curr}, \varepsilon)$} \label{alg:gen_model}
        \State{$c = \textit{coverageCheck}(h, \widehat{\mathcal{I}}, \varepsilon)$} \label{alg:cov}
        \If{$c < \tau_{\text{cov}}$}
            \State{\textbf{break}} \label{alg:skip}
        \EndIf
        \State{$\mathcal{H} = \mathcal{H} \cup \{(h,\widehat{\mathcal{I}})\}$} \label{alg:add_pool}
        \State{$\mathcal{D}_{curr} = \mathcal{D}_{curr} \setminus \textit{dilate}(\widehat{\mathcal{I}}, \beta\varepsilon)$} \label{alg:buffer}
    \EndWhile
    \State{$\mathcal{D}_{curr} = \mathcal{D}$} \label{alg:reset}
\EndFor \label{alg:run_end}
\State{$\{\theta_k\}_{k=1}^{\kappa} = \textit{solveMaxCoverage}(\textcolor{black}{\mathcal{H}}, \kappa)$} \label{alg:maxcov}
\State{\textbf{return} $\{\theta_k\}_{k=1}^{\kappa}$}
\end{algorithmic}
\end{algorithm}

\section{Experiments}

We evaluate the proposed geometric-aware inlier refinement (GAIR) by analyzing its behavior in both single-model superquadric fitting and multi-model primitive decomposition scenarios. Specifically, we assess its integration within our GAIR-\ransac and GAIR-\ransacov on synthetic and realistic point clouds under controlled noise and outlier conditions. 
\textcolor{black}{The code is publicly available at
\url{https://github.com/Tededo02/3D_superquadric_decomposition/tree/gair_ransac}}.\\

\subsection{Datasets}
\label{sec:dataset}
\begin{figure*}
\centering
\includegraphics[width=\linewidth]{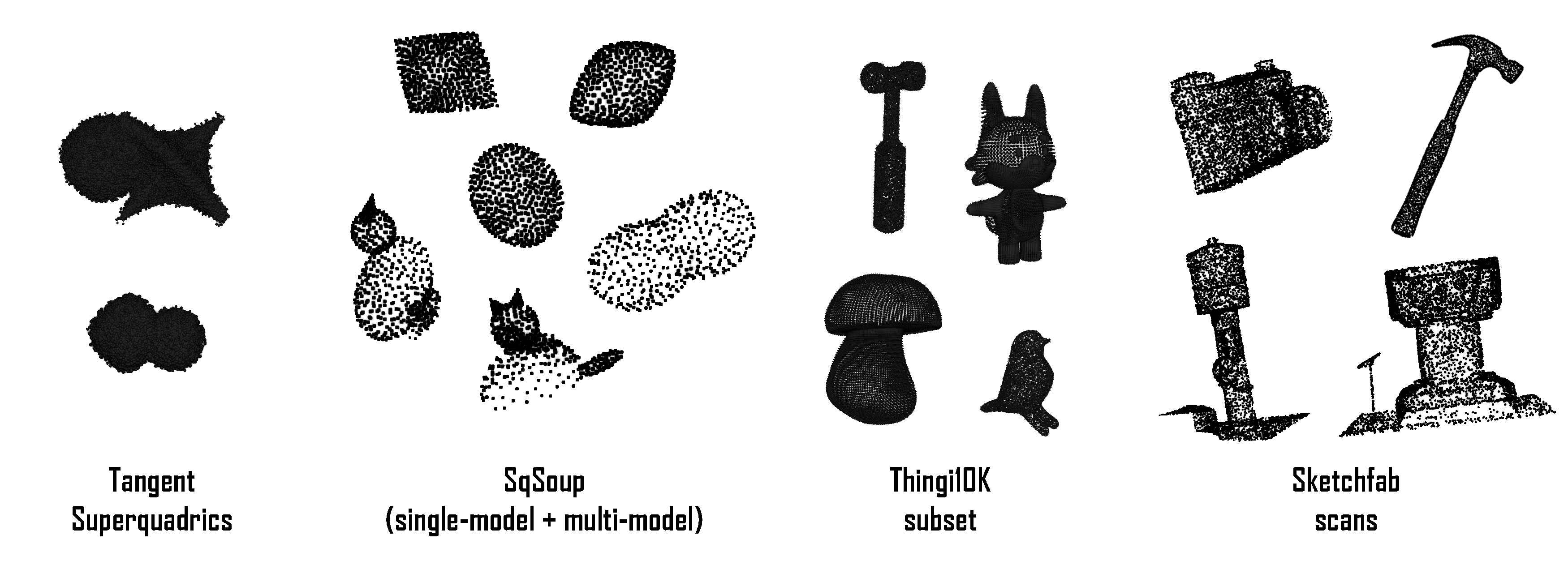}
\caption{
Overview of the datasets used in our experiments.
From left to right: synthetic \texttt{TangentSuperquadrics} point clouds, used to
evaluate pure fitting accuracy; \texttt{SqSoup} in both the single-model and multi-model configuration, used to test the algorithm in a setup with few points; a subset of CAD-like shapes
from \texttt{Thingi10K}, showcasing the applicability of our method to more complex and realistic geometries, and, finally, a set of four 3d scans from \textit{Sketchfab}. We also considered real LiDAR data  reported in Fig. \ref{fig:real_pc}
}
\label{fig:dataset1}
\end{figure*}

For our evaluation, we consider three complementary types of data:
(i) controlled synthetic benchmarks, 
(ii) challenging synthetic configurations, 
and (iii) real-world shapes. An overview is reported in Figure~\ref{fig:dataset1}.
\textcolor{black}{All shapes are normalized to a common scale to ensure that evaluation metrics, such as the Chamfer distance and Hausdorff distance, are comparable.}

\textbf{\texttt{SqSoup} (synthetic).}
The \texttt{SqSoup} dataset is a collection of synthetic point clouds specifically designed for superquadric fitting evaluation. 
We consider two subsets: a \texttt{SqSoup} \texttt{single}-\texttt{model} split, containing point clouds sampled from individual superquadrics with varying shape parameters, and a \texttt{SqSoup multi-model} split, comprising composite scenes constructed from multiple interacting superquadrics (e.g., cat-like shapes and a peanut). 
These multi-model scenes are particularly useful for stress-testing primitive decomposition algorithms, as the constituent primitives exhibit partial overlap and ambiguous spatial configurations.

\textbf{\texttt{TangentSuperquadrics.}}
We introduce synthetic scenes where multiple superquadrics are placed close to each other.
These configurations are particularly challenging for primitive decomposition, as adjacent
primitives exhibit minimal spatial separation while remaining geometrically distinct.

In such cases, methods relying solely on point-to-model distance or spatial proximity tend to
incorrectly merge neighboring primitives into a single model. These examples therefore act as a
stress-test for the inlier refinement step.

\textbf{\texttt{Thingi10K} subset (\texttt{3D shapes}).}
To assess generalization beyond ideal superquadric configurations, we include four shapes from the \texttt{Thingi10K} repository: a mushroom, a bird, a hammer, and an anthropomorphic cartoon character.
These models exhibit irregular geometry, non-convex surfaces, and structures that only loosely resemble superquadric primitives, providing a more realistic evaluation scenario.

\textcolor{black}{
\textbf{\texttt{Sketchfab scans}.}
To evaluate GAIR beyond synthetic and CAD-like benchmarks, we
consider four publicly available real-world 3D scans downloaded from
Sketchfab: a baptismal font~\cite{sketchfab_baptismal_font}, a fire
hydrant~\cite{sketchfab_hydrant}, a film
camera~\cite{sketchfab_camera}, and a
hammer~\cite{sketchfab_hammer}. The models were acquired using
either photogrammetry or dedicated 3D scanning equipment. Since
ground-truth primitive decompositions are not available, performance
on these data is assessed qualitatively. The downloaded meshes were
converted into point clouds by uniformly subsampling
8000 surface points. 
}

\textcolor{black}{
\textbf{\texttt{LiDAR scans} (real data).}
We additionally acquired three point-cloud scenes using the LiDAR
sensor of an iPhone 16 Pro using PointCloud Scanner: a ball placed on top of a box, a sofa with
cushions, and a plush toy. The three scenes provide increasing levels
of difficulty. The ball-and-box scene represents a simple controlled
configuration composed of two clearly separated primitives. The sofa
scene contains adjacent and partially overlapping structures, with
limited visibility of some cushions. Finally, the plush-toy scan is
characterized by a noisy fuzzy surface and weak geometric separation
between its constituent parts.
}

\subsection{Metrics}\label{sec:metrics}

We evaluate the performance of the proposed method along
\textcolor{black}{three} complementary dimensions:
\textcolor{black}{inlier-outlier classification accuracy}, geometric accuracy,
\textcolor{black}{and} computational efficiency.

\textcolor{black}{
Inlier-outlier classification performance is measured using the
\emph{Inlier Assignment Error} (IAE), which quantifies the fraction of
points incorrectly classified as either inliers or outliers. Points
mi belonging to the target models are labeled as inliers, whereas the
explicitly introduced outlier points are labeled as outliers. Let
$y_i \in \{0,1\}$ denote the ground-truth binary label of point
$\mathbf{x}_i$, and let $\hat{y}_i$ denote its predicted label, where
$1$ indicates an inlier and $0$ an outlier. The IAE is defined as:}
{\color{black}
\begin{equation}
    \mathrm{IAE}
    =
    \frac{1}{N}
    \sum_{i=1}^{N}
    \mathbb{I}\!\left[\hat{y}_i \neq y_i\right],
    \label{eq:iae}
\end{equation}}\textcolor{black}{where lower values indicate better classification performance.
Therefore, the IAE enables a consistent
evaluation of inlier--outlier classification across all datasets
without requiring each point to be associated with a specific
ground-truth primitive.
}

Geometric fidelity is assessed using the \emph{Chamfer Distance} (CD),
which captures the average discrepancy between the reconstructed and
ground-truth surfaces, and the \emph{Hausdorff Distance} (HD), which
reflects the worst-case deviation.
\textcolor{black}{
For the \texttt{Thingi10K}, \texttt{SqSoup} datasets, we report the
\emph{Normalized Chamfer Distance} and the
\emph{Normalized Hausdorff Distance}, computed after normalizing the
point-cloud bounding boxes to ensure that metric values are comparable
across shapes within the same experiment.
}

Finally, we report \emph{execution time} and the number of local
optimization steps (\emph{convergence}), providing insight into the
computational cost and \textcolor{black}{behavior} of the different methods.

\subsection{Baselines and evaluation protocols}
We compare the proposed method against three well-established variants of the \ransac paradigm, which can be interpreted as progressively enriching the inlier refinement process.

Vanilla \ransac~\cite{RANSAC} represents the baseline formulation, where inliers are selected solely based on a residual threshold, without any refinement. 
LO-\ransac~\cite{LO} introduces a local optimization step that refines the inlier set using residual-based criteria, improving model estimation without incorporating additional structural information. 
GC-\ransac~\cite{GCRANSAC} further extends this formulation by introducing spatial coherence through a graph-cut optimization, promoting inlier assignments that are locally consistent in Euclidean space.

Within this framework, the proposed method can be interpreted as extending inlier refinement with geometric priors, by incorporating normal consistency in addition to residual and spatial information.

For the multi-model case, we adopt a sequential strategy: all methods are embedded into a Sequential-\textsc{RANSAC} framework~\cite{seqran}, where models are extracted one at a time and their inliers removed before reapplying the algorithm to the residual data. This procedure allows us to evaluate the robustness of each variant in primitive decomposition tasks. 

In the final experiment, we further improve the decomposition quality by
applying a  \ransacov selection step~\cite{magri2016multiple} as post-processing.
After the sequential procedure has terminated, all candidate models collected
across the $n$ rounds are passed to  \ransacov , which solves an Integer Linear
Program to select the subset of at most $\kappa$ models that jointly maximise the
number of explained inliers.

To ensure a fair comparison, we configure all methods under the same conditions:
The maximum number of iterations $m$ is fixed to $20$ for all algorithms. A larger budget would eventually allow all methods to saturate, thus hiding meaningful performance differences.
The consensus function is the standard count of inliers.
The sample-set size $\lvert\mathcal{M}_j\rvert$ was set to 30 by default. For larger point clouds in the uncapped experiments, however, the sample-set size was increased proportionally to the number of points in the point cloud.

Non-linear model fitting for superquadrics is performed using \texttt{scipy.optimize.least\_squares} solver,
{\color{black}using a trust-region reflective method, PCA-based initialization, bounded parameters, and a robust soft-$\ell_1$ loss.}
For algorithms involving an inner \textsc{RANSAC} loop (GC-\textsc{RANSAC} and GAIR-\textsc{RANSAC}), we fix the number of inner iterations to 25.
This setup allows us to isolate the contribution of the refinement strategies and proposed energy formulation, while keeping all other factors consistent across the baselines. \textcolor{black}{The default parameter values used in the experimental evaluation are reported in Table~\ref{tab:hyperparameters}.
}

\begin{table}[t]
\centering
\caption{Default parameters used in the experiments.}
\label{tab:hyperparameters}

\footnotesize
\setlength{\tabcolsep}{2.5pt}
\renewcommand{\arraystretch}{1.05}

\begin{tabular}{
    @{}
    p{0.20\columnwidth}
    p{0.53\columnwidth}
    c
    @{}
}
\toprule
\textbf{Component} & \textbf{Parameter} & \textbf{Value} \\
\midrule
\ransac
    & Inlier threshold, synthetic data
    & $2.5\sigma$ \\

    & Inlier threshold, real scans
    & $0.015\,\mathrm{m}$ \\

    & Outer iterations
    & $20$ \\
\midrule
Local opt.
    & Inner iterations
    & $25$ \\
\midrule
Graph
    & Neighbors, controlled data
    & $k=6$ \\

    & Neighbors, real scans
    & $k=10$ \\
\midrule
Hypothesis generation
    & Sample size
    & $30$ \\

    & Minimum inlier support
    & $20$ \\

    & Minimum surface coverage
    & $\tau_{\mathrm{cov}}=0.4$ \\
\bottomrule
\end{tabular}
\end{table}

\subsection{Sensitivity to the inlier threshold}
\label{sec:thresh}
\begin{table}[]
\centering
\caption{Sensitivity to the inlier threshold $\varepsilon = s\sigma$ on \texttt{TangentSuperquadrics}.
Chamfer Distance (CD), Hausdorff Distance (HD) averaged over runs.
Best values in bold.}
\label{tab:epsilon}
\resizebox{\columnwidth}{!}{%
\begin{tabular}{llccc}
\toprule
Scale $s$ & Method & CD $\downarrow$ & HD $\downarrow$ \\
\midrule

\multirow{2}{*}{$1.0$}
 & Vanilla \ransac
 & 0.64$\pm$0.10
 & 2.15$\pm$0.45\\
 & GAIR-\ransac (ours)
 & \textbf{0.50}$\pm$0.01
 & \textbf{1.73}$\pm$0.28\\

\midrule
\multirow{2}{*}{$1.5$}
 & Vanilla \ransac
 & 0.70$\pm$0.07
 & 2.06$\pm$0.22\\
 & GAIR-\ransac (ours)
 & \textbf{0.43}$\pm$0.04
 & \textbf{0.96}$\pm$0.16\\

\midrule
\multirow{2}{*}{$2.0$}
 & Vanilla \ransac
 & 0.67$\pm$0.07
 & 2.01$\pm$0.10\\
 & GAIR-\ransac (ours)
 & \textbf{0.44}$\pm$0.02
 & \textbf{1.62}$\pm$0.21\\

\midrule
\multirow{2}{*}{$2.5$}
 & Vanilla \ransac
 & 0.65$\pm$0.10
 & 2.22$\pm$0.19\\
 & GAIR-\ransac (ours)
 & \textbf{0.41}$\pm$0.01
 & \textbf{1.30}$\pm$0.18\\

\midrule
\multirow{2}{*}{$3.0$}
 & Vanilla \ransac
 & 0.68$\pm$0.03
 & 2.54$\pm$0.14\\
 & GAIR-\ransac (ours)
 & \textbf{0.43}$\pm$0.03
 & \textbf{1.42}$\pm$0.36\\

\midrule
\multirow{2}{*}{$3.5$}
 & Vanilla \ransac
 & 0.70$\pm$0.15
 & 2.43$\pm$0.67\\
 & GAIR-\ransac (ours)
 & \textbf{0.42}$\pm$0.01
 & \textbf{1.58}$\pm$0.09\\

\midrule
\multirow{2}{*}{$4.0$}
 & Vanilla \ransac
 & 0.61$\pm$0.06
 & 2.33$\pm$0.55\\
 & GAIR-\ransac (ours)
 & \textbf{0.42}$\pm$0.01
 & \textbf{1.56}$\pm$0.09\\

\bottomrule
\end{tabular}}
\end{table}

We first analyze the sensitivity of our method to the inlier threshold parameter $\varepsilon$, which plays a critical role in consensus maximization frameworks. 
While all methods rely on this parameter to determine inlier membership, we show that geometric-aware refinement reduces its impact, leading to more stable inlier sets even when the threshold is not perfectly tuned. 

To this end, we compare GAIR-\ransac with Vanilla-\ransac on the \texttt{TangentSuperquadrics}. 
Each point cloud is corrupted with Gaussian noise of fixed standard deviation $\sigma = 0.4$. 
Additionally, $10\%$ of the points are replaced with uniformly distributed outliers within the bounding box of the point cloud.  
We vary the inlier threshold as $\varepsilon = s \cdot \sigma$, with $s \in [1, 4]$ with step $0.5$ and report Chamfer Distance (CD) and Hausdorff Distance (HD), averaged over 5 runs per configuration in Table~\ref{tab:epsilon}. 
Two main trends can be observed. First, GAIR-\ransac consistently outperforms Vanilla \ransac across all values of $s$, confirming the benefit of geometric-aware refinement independently of the threshold choice. 
Second, and more importantly, GAIR-\ransac exhibits significantly lower sensitivity to the threshold parameter: performance remains stable over a wide range of values, with only minor variations in CD and HD. 
In contrast, Vanilla \ransac is strongly affected by the choice of $s$. 
Small values ($s \leq 2$) lead to an underestimation of the inlier set and degraded model quality, while large values ($s \geq 3.5$) admit points far from the surface, increasing fitting error. 
This behaviour reflects the fact that, in proximity-based formulations, the threshold alone governs inlier selection.

The improved robustness of GAIR-\ransac can be attributed to the additional geometric constraints introduced in the refinement step. 
By enforcing normal consistency, inlier selection is no longer driven solely by point-to-model distance, but also by local surface coherence. 
As a result, the method is less dependent on the precise tuning of $\varepsilon$, making the threshold easier to set in practice.

Based on this analysis, we fix $\varepsilon = 2.5\,\sigma$ for all subsequent \textcolor{black}{ controlled synthetic} experiments, which provides a good trade-off between geometric accuracy and robustness.

\textcolor{black}{In practice, for acquired point clouds, the noise level $\sigma$ depends on the acquisition process, including the measurement accuracy of the sensing device and the physical units in which the scan is represented. In these cases, the threshold $\varepsilon$ has a direct geometric interpretation, as it corresponds to the maximum point-to-surface discrepancy that is considered compatible. It can therefore be selected according to the nominal accuracy of the acquisition device or to an externally estimated noise level $\sigma$. Automatically estimating  $\sigma$ from the input data constitutes a separate, sensor-dependent problem, which is outside the scope of the present contribution. Moreover, our sensitivity analysis shows that GAIR remains stable over a relatively broad range of threshold values. For this reason, in the real-data experiments we use a single fixed threshold for each dataset, expressed in the physical units of the corresponding scans as reported in Tab.\ref{tab:hyperparameters}.
}

\subsection{Single-model fitting}
\begin{table}[t]
\centering
\caption{Single-model fitting: mean \textcolor{black}{IAE} across the three single model point clouds from \emph{SqSoup} and 25 trials per condition. Best result per row in \textbf{bold}.}
\label{tab:single_model_iae}
\begin{tabular}{lcccc}
\toprule
Outlier & GAIR & GC & LO & \ransac \\
\midrule
0\%  & \textbf{0.0260} & 0.0478 & 0.0480 & 0.0595 \\
5\%  & \textbf{0.0302} & 0.0569 & 0.0590 & 0.0823 \\
10\% & \textbf{0.0226} & 0.0493 & 0.0522 & 0.1113 \\
15\% & \textbf{0.0302} & 0.0583 & 0.0597 & 0.1143 \\
20\% & \textbf{0.0318} & 0.0594 & 0.0701 & 0.1465 \\
25\% & \textbf{0.0361} & 0.0700 & 0.0754 & 0.1425 \\
30\% & \textbf{0.0387} & 0.0739 & 0.0919 & 0.1811 \\
35\% & \textbf{0.0441} & 0.0757 & 0.1011 & 0.2122 \\
40\% & \textbf{0.0370} & 0.0725 & 0.1091 & 0.2295 \\
\bottomrule
\end{tabular}
\end{table}
We evaluate the effect of inlier refinement in the single-model setting, where the goal is to isolate improvements in fitting accuracy independently of segmentation ambiguities. In this controlled scenario, we consider the single model split of the \texttt{SqSoup} dataset. Each point cloud contains a single superquadric corrupted by increasing levels of uniformly distributed outliers, allowing us to focus purely on the robustness of the fitting process.

\begin{figure}[]
    \centering
    \includegraphics[width=\linewidth]{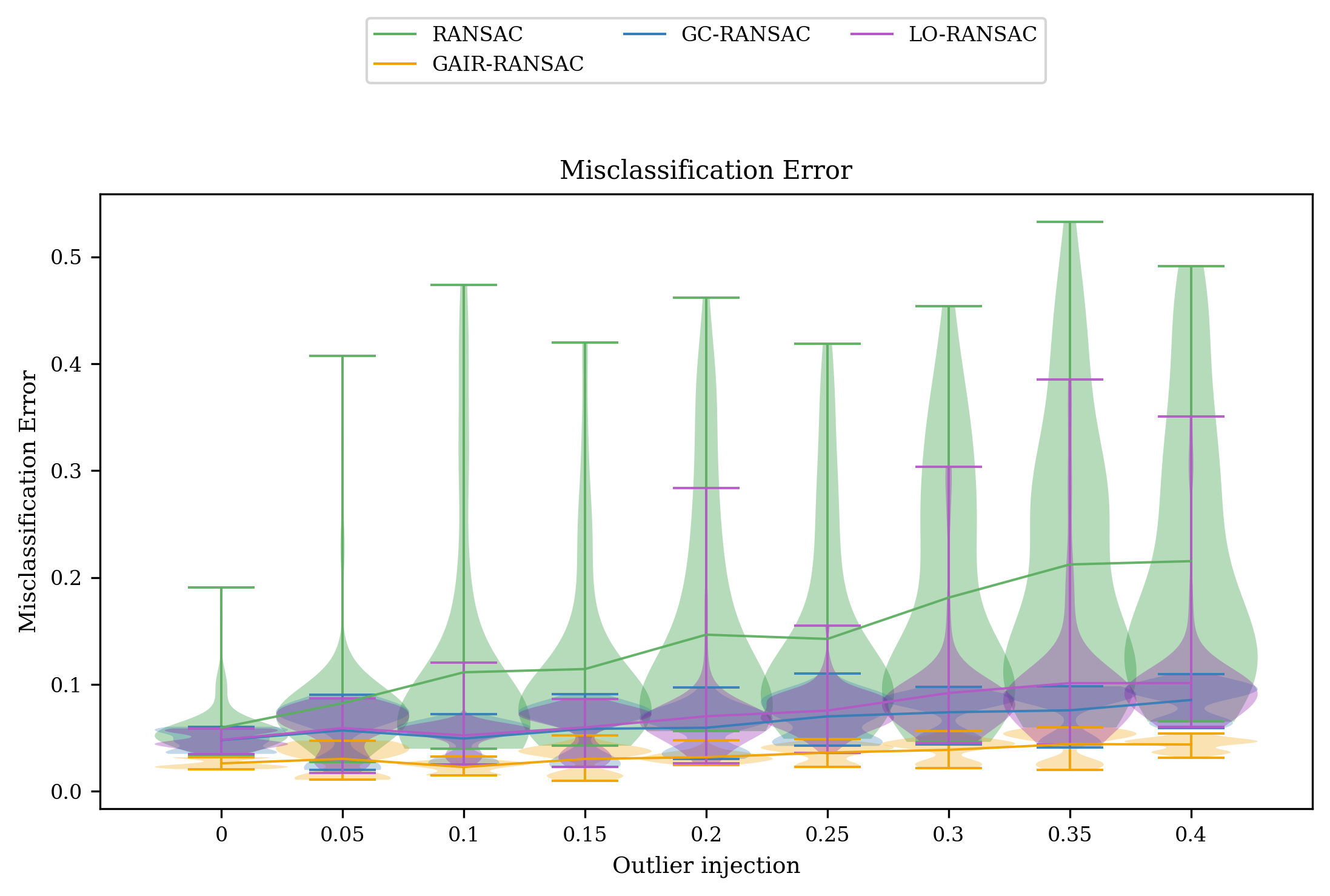}
    \caption{Average \textcolor{black}{IAE} as a function of the outlier ratio on the single-model subset of \emph{SqSoup}.}
    \label{fig:single_model_me_outlier}
\end{figure}

Results are reported in Table~\ref{tab:single_model_iae} and in Figure \ref{fig:single_model_me_outlier}. As expected, the performance of all methods degrades as the outlier ratio increases; however, clear trends emerge across refinement strategies.
First, Vanilla \ransac, which relies solely on residual-based inlier selection, consistently achieves the worst performance. Its error rapidly increases with the outlier ratio, exceeding 20\% \textcolor{black}{IAE} at high contamination levels, confirming the limitations of purely residual-based criteria in the presence of significant noise.
Introducing local optimization (LO-\ransac) improves the results, but only marginally: while it reduces the error at low outlier ratios, its performance still deteriorates significantly as contamination increases, roughly doubling between 0\% and 40\% outliers. This indicates that residual-based refinement alone is insufficient to ensure robustness.
A more substantial improvement is observed with GC-\ransac, where spatial coherence is enforced through proximity-based regularization. This leads to more stable behavior across outlier levels, highlighting the benefit of incorporating additional priors beyond point-wise residuals.

Finally, GAIR-\ransac consistently achieves the best performance across all contamination levels. Even in this simplified setting, where no interaction between primitives is present, the geometric-aware refinement provides a clear advantage, reducing the \textcolor{black}{IAE} by approximately a factor of two compared to GC-\ransac. This confirms that incorporating geometric consistency, in particular normal coherence, leads to more reliable inlier selection and significantly improves robustness to outliers.

\begin{figure}[]
    \centering
    \includegraphics[width=\linewidth]{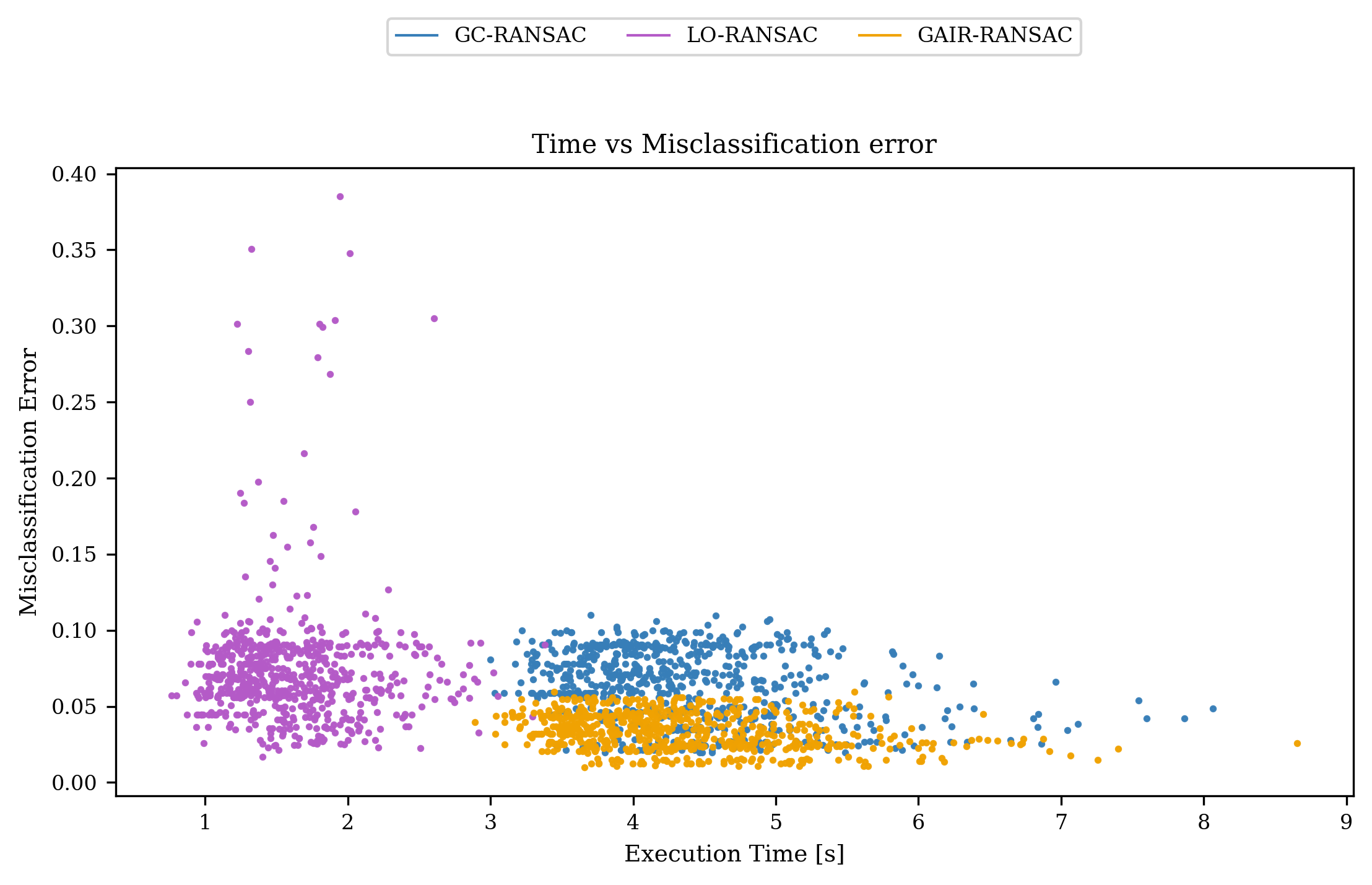}
    \caption{\textcolor{black}{IAE} versus execution time for the single-model subset of \emph{SqSoup}.}
    \label{fig:single_model_me_time}
\end{figure}

In addition to accuracy, we analyze the trade-off between \textcolor{black}{IAE} and computational cost in Figure~\ref{fig:single_model_me_time}, which reports the results of multiple runs for GAIR-\ransac, GC-\ransac, and LO-\ransac. Each point corresponds to a single run, allowing us to visualize both performance and variability.
LO-\ransac exhibits a lower computational cost, but at the price of significantly higher \textcolor{black}{IAE}. In contrast, GC-\ransac and GAIR-\ransac show comparable execution times, reflecting the additional cost of the graph-cut-based refinement. However, GAIR-\ransac consistently achieves lower error across runs, clearly dominating GC-\ransac in terms of the accuracy–efficiency trade-off.

\subsection{Primitive decomposition}
We now move to the multi-model setting, where the goal is to decompose a point cloud into multiple primitives. In contrast to the single-model case, this scenario introduces intrinsic ambiguities, as different primitives may be spatially adjacent or even intersect, making inlier assignment significantly more challenging.
To address this setting, we extend all methods from their single-model \ransac formulation to a multi-model framework based on \ransacov, which decouples hypothesis generation from model selection, as described in Section~\ref{sec:gair_ransacov}, while differing in their inlier refinement strategies.

\paragraph{Tangent superquadrics.}
\begin{table}[t]

\centering
\caption{Primitive decomposition on the synthetic scene \emph{tangent superquadrics} consisting of $\kappa=4$ intertwined superquadrics.
 Best values in \textbf{bold}.}
\label{tab:primitive_decomposition}
\resizebox{\columnwidth}{!}{%
\begin{tabular}{llccc}
\toprule
$(\sigma,\, N_{\text{out}})$ & Method & CD $\downarrow$ & HD $\downarrow$ & \textcolor{black}{IAE} $\downarrow$ \\
\midrule
\multirow{2}{*}{$(0.1,\; 0)$}
 & GC-\ransacov           & 0.82$\pm$1.07 & 5.09$\pm$8.87 & 0.06$\pm$0.06 \\
 & GAIR-\ransacov  (ours)  & \textbf{0.29}$\pm$0.01 & \textbf{0.58}$\pm$0.03 & \textbf{0.03}$\pm$0.00 \\
\midrule
\multirow{2}{*}{$(0.1,\; 4000)$}
 & GC-\ransacov            & 0.38$\pm$0.20 & 1.59$\pm$1.91 & 0.08$\pm$0.09 \\
 & GAIR-\ransacov  (ours)  & \textbf{0.27}$\pm$0.00 & \textbf{0.59}$\pm$0.12 & \textbf{0.03}$\pm$0.00 \\
\midrule
\multirow{2}{*}{$(0.2,\; 0)$}
 & GC-\ransacov            & 0.29$\pm$0.01 & 1.51$\pm$0.49 & 0.09$\pm$0.00 \\
 & GAIR-\ransacov  (ours)  & \textbf{0.26}$\pm$0.01 & \textbf{0.91}$\pm$0.20 & \textbf{0.08}$\pm$0.00 \\
\midrule
\multirow{2}{*}{$(0.2,\; 4000)$}
 & GC-\ransacov            & 0.28$\pm$0.01 & 1.29$\pm$0.39 & 0.08$\pm$0.00 \\
 & GAIR-\ransacov  (ours)  & \textbf{0.26}$\pm$0.00 & \textbf{1.10}$\pm$0.12 & \textbf{0.08}$\pm$0.00 \\
\midrule
\multirow{2}{*}{$(0.4,\; 0)$}
 & GC-\ransacov            & 0.70$\pm$0.18 & 3.22$\pm$1.43 & 0.14$\pm$0.04 \\
 & GAIR-\ransacov  (ours)  & \textbf{0.63}$\pm$0.00 & \textbf{2.05}$\pm$0.12 & \textbf{0.09}$\pm$0.00 \\
\midrule
\multirow{2}{*}{$(0.4,\; 4000)$}
 & GC-\ransacov            & 0.59$\pm$0.12 & 2.97$\pm$1.07 & 0.11$\pm$0.03 \\
 & GAIR-\ransacov  (ours)  & \textbf{0.55}$\pm$0.12 & \textbf{1.98}$\pm$0.22 & \textbf{0.09}$\pm$0.01 \\
\bottomrule
\end{tabular}}
\end{table}

We first evaluate the two best-performing methods from the previous experiment, namely GC-\ransacov and GAIR-\ransacov, on the \texttt{Tangent} \texttt{Superquadrics} dataset introduced in Section~\ref{sec:dataset}. This benchmark is specifically designed to stress-test primitive decomposition in the presence of adjacent and touching surfaces.

Using the threshold scale $s = 2.5$ identified in Section~\ref{sec:thresh}, we compare the two methods under varying levels of Gaussian noise $\sigma$ and injected outliers $N_{\mathrm{out}}$. In this setting, spatial proximity becomes an unreliable cue, as points belonging to different primitives can lie arbitrarily close in Euclidean space.

Performance is evaluated in terms of Chamfer Distance (CD) and Hausdorff Distance (HD), which quantify how accurately the underlying geometry is recovered, as well as \textcolor{black}{Inlier Assignment Error (IAE)}. Results reported in Table~\ref{tab:primitive_decomposition} show that GAIR-\ransac consistently outperforms GC-\ransac across all conditions and metrics.

The improvement is systematic: GAIR achieves lower CD and HD, indicating a more faithful geometric reconstruction, and higher segmentation accuracy, reflecting a more reliable separation of adjacent primitives. Moreover, the variance of the results is consistently lower, highlighting the increased stability of the geometric-aware refinement. These findings confirm that incorporating geometric priors, in particular normal consistency, is crucial in scenarios where proximity alone is insufficient to disambiguate surface membership.
\begin{figure*}
    \centering
    \begin{subfigure}[t]{0.48\linewidth}
        \centering
        \includegraphics[width=\linewidth]{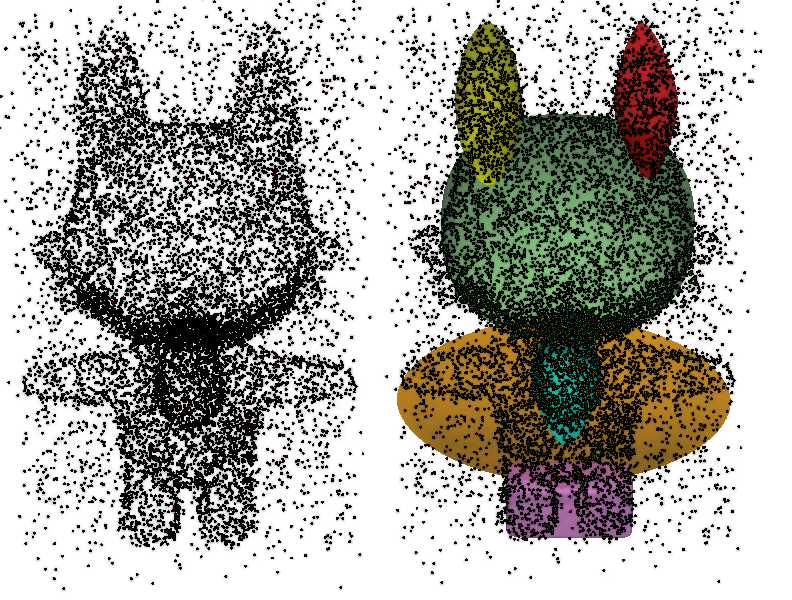}
        \caption{Cartoon Character (high outliers)}
        \label{fig:robustness}
    \end{subfigure}
    \hfill
    \begin{subfigure}[t]{0.48\linewidth}
        \centering
        \includegraphics[width=\linewidth]{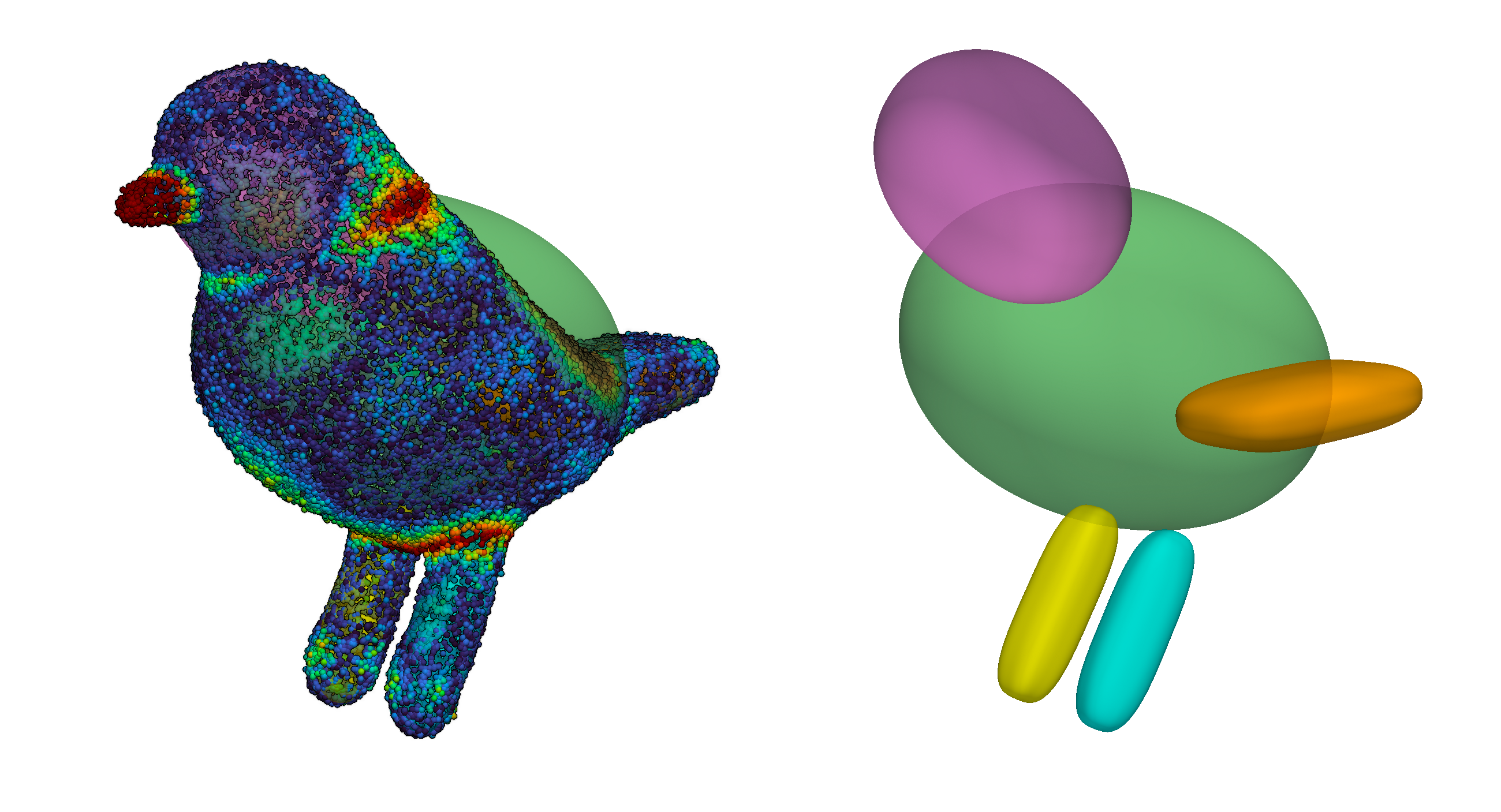}
        \caption{Bird}
        \label{fig:bird}
    \end{subfigure}
    
    \caption{
    Qualitative results of primitive decomposition using GAIR-\ransacov on shapes from the \emph{3D Shapes} dataset.
    (a) Robustness under severe outlier contamination: despite strong noise, the decomposition remains consistent with the underlying geometry, as normal coherence prevents incorrect inlier assignments.
    (b) Example on a bird point cloud, illustrating the ability of the method to recover meaningful primitives on complex shapes.
    }
    \label{fig:qualitative_results}
\end{figure*}

\begin{table}[t]
\centering
\caption{\textcolor{black}{Chamfer distance (CD) and Inlier Assignment Error (IAE) as a function of the injected outlier fraction, averaged across point clouds.}}
\label{tab:outlier_results}

\textbf{Chamfer Distance $\downarrow$} \\[2pt]
\begin{tabular}{lcccc}
\toprule
Outlier Frac. & GAIR & GC & LO & Vanilla \\
\midrule
0.00 & \textbf{0.0477} & 0.0789 & 0.0847 & 0.0851 \\
0.05 & \textbf{0.0543} & 0.0879 & 0.0956 & 0.1030 \\
0.10 & \textbf{0.0592} & 0.0908 & 0.1065 & 0.1133 \\
0.15 & \textbf{0.0659} & 0.0971 & 0.1109 & 0.1117 \\
0.20 & \textbf{0.0657} & 0.1051 & 0.1158 & 0.1215 \\
0.25 & \textbf{0.0687} & 0.1031 & 0.1224 & 0.1275 \\
0.30 & \textbf{0.0726} & 0.1105 & 0.1304 & 0.1323 \\
0.35 & \textbf{0.0753} & 0.1123 & 0.1349 & 0.1390 \\
0.40 & \textbf{0.0771} & 0.1100 & 0.1470 & 0.1464 \\
\bottomrule
\end{tabular}

\vspace{8pt}
\textbf{Inlier Assignment Error (IAE) $\downarrow$} \\[2pt]
\begin{tabular}{lcccc}
\toprule
Outlier Frac. & GAIR & GC & LO & Vanilla \\
\midrule
0.00 & 0.0604 & \textbf{0.0514} & 0.0668 & 0.0755 \\
0.05 & 0.0599 & \textbf{0.0635} & 0.0914 & 0.1002 \\
0.10 & \textbf{0.0663} & 0.0896 & 0.1214 & 0.1341 \\
0.15 & \textbf{0.0765} & 0.1055 & 0.1457 & 0.1571 \\
0.20 & \textbf{0.0804} & 0.1464 & 0.1717 & 0.1819 \\
0.25 & \textbf{0.0853} & 0.1610 & 0.1966 & 0.2054 \\
0.30 & \textbf{0.0975} & 0.1872 & 0.2219 & 0.2259 \\
0.35 & \textbf{0.1081} & 0.2124 & 0.2488 & 0.2540 \\
0.40 & \textbf{0.1141} & 0.2354 & 0.2693 & 0.2786 \\
\bottomrule
\end{tabular}
\end{table}
\begin{table}
\caption{Number of local optimization steps and runtime as a function of the injected outlier fraction, averaged across point clouds.}
\label{tab:outlier_runtime}
\vspace{8pt}
\textbf{Number of Local Optimizations} \\[2pt]
\begin{tabular}{lcccc}
\toprule
Outlier Frac. & GAIR & GC & LO & Vanilla \\
\midrule
0.00 & 19.10 & 16.22 & 4.34 & 0.00 \\
0.05 & 18.00 & 16.08 & 4.68 & 0.00 \\
0.10 & 23.12 & 18.55 & 5.23 & 0.00 \\
0.15 & 24.90 & 20.29 & 5.31 & 0.00 \\
0.20 & 25.49 & 21.42 & 5.94 & 0.00 \\
0.25 & 26.10 & 22.60 & 5.70 & 0.00 \\
0.30 & 27.20 & 22.26 & 6.30 & 0.00 \\
0.35 & 28.95 & 23.66 & 6.92 & 0.00 \\
0.40 & 28.29 & 21.88 & 7.13 & 0.00 \\
\bottomrule
\end{tabular}

\vspace{8pt}
\textbf{Runtime (s)} \\[2pt]
\begin{tabular}{lcccc}
\toprule
Outlier Frac. & GAIR & GC & LO & Vanilla \\
\midrule
0.00 & 52.33  & 58.15  & 21.05 & 14.64 \\
0.05 & 52.91  & 56.95  & 26.23 & 20.78 \\
0.10 & 63.16  & 71.23  & 28.39 & 22.61 \\
0.15 & 70.36  & 78.92  & 32.66 & 27.45 \\
0.20 & 72.38  & 86.08  & 36.59 & 30.17 \\
0.25 & 70.73  & 90.07  & 35.95 & 31.54 \\
0.30 & 77.51  & 97.43  & 39.52 & 33.73 \\
0.35 & 83.23  & 104.96 & 44.44 & 36.03 \\
0.40 & 82.05  & 100.12 & 45.25 & 38.29 \\
\bottomrule
\end{tabular}
\end{table}

\paragraph{Primitive decomposition on \texttt{SqSoup} and \texttt{3D Shapes}.}

\begin{figure*}[]
    \centering
    \begin{subfigure}[t]{0.32\linewidth}
        \centering
        \includegraphics[width=\linewidth]{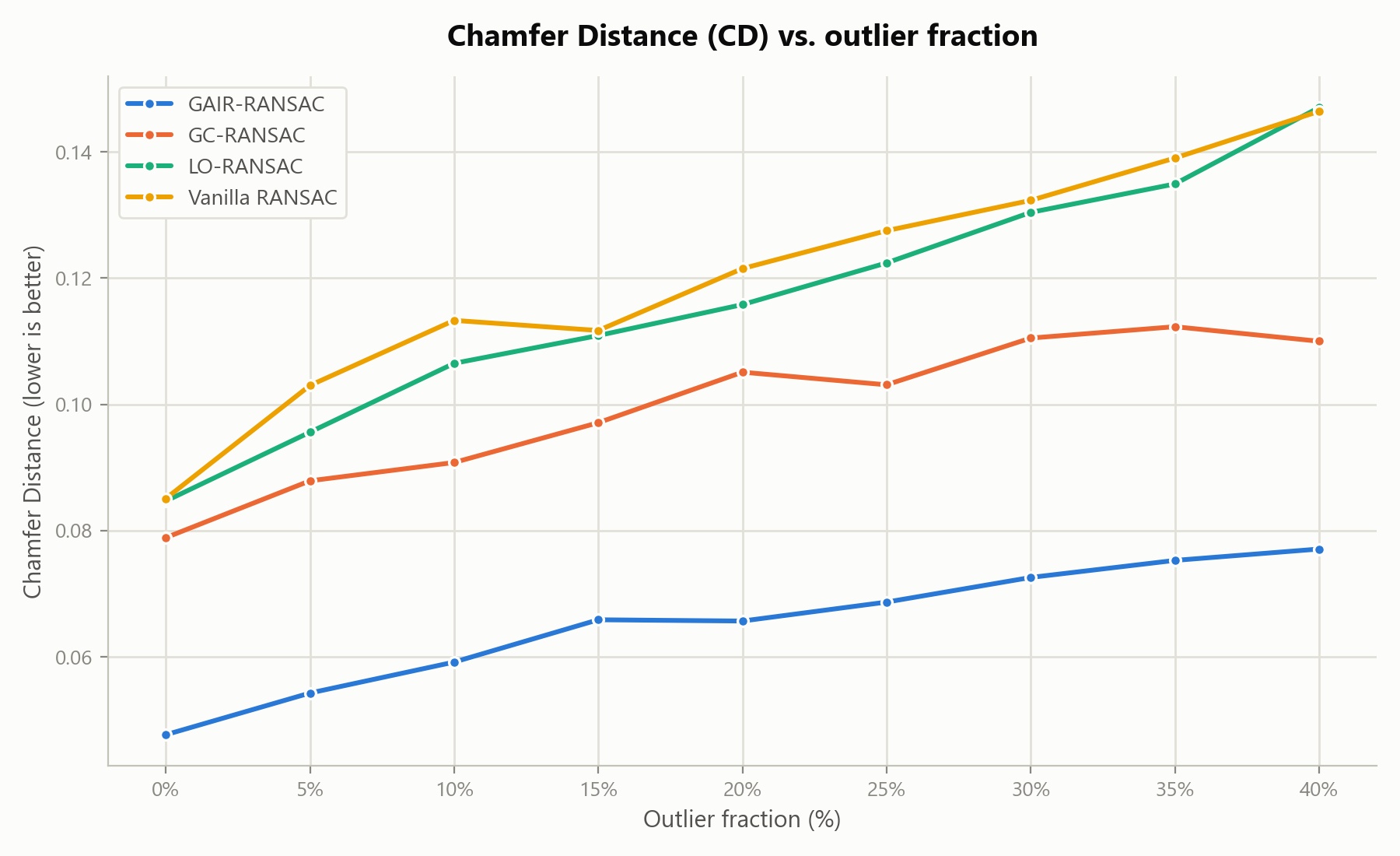}
        \caption{Chamfer Distance (CD)}
        \label{fig:plotme1}
    \end{subfigure}
    \hfill
    \begin{subfigure}[t]{0.32\linewidth}
        \centering
        \includegraphics[width=\linewidth]{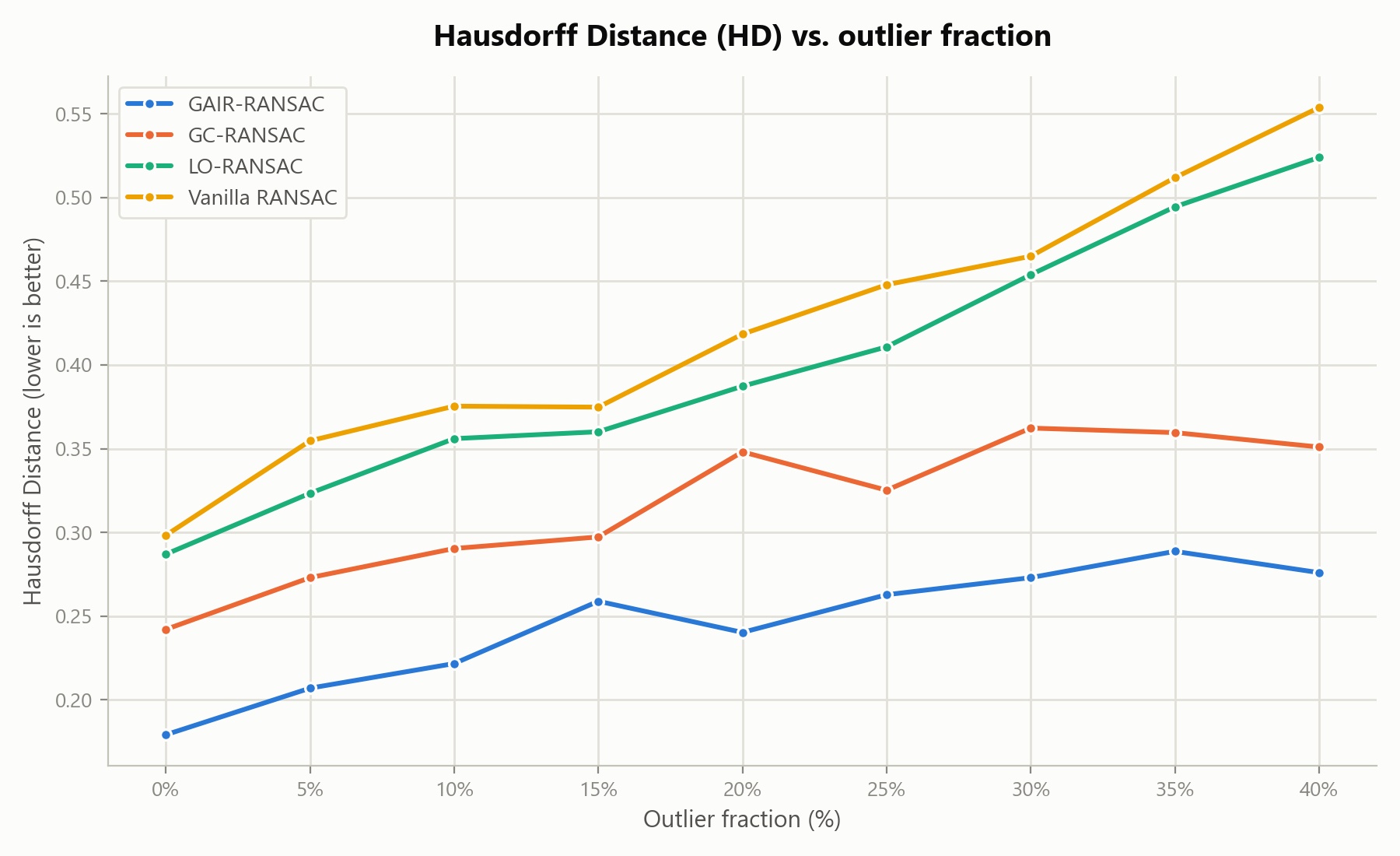}
        \caption{Hausdorff Distance (HD)}
        \label{fig:plotme2}
    \end{subfigure}
    \hfill
    \begin{subfigure}[t]{0.32\linewidth}
        \centering
        \includegraphics[width=\linewidth]{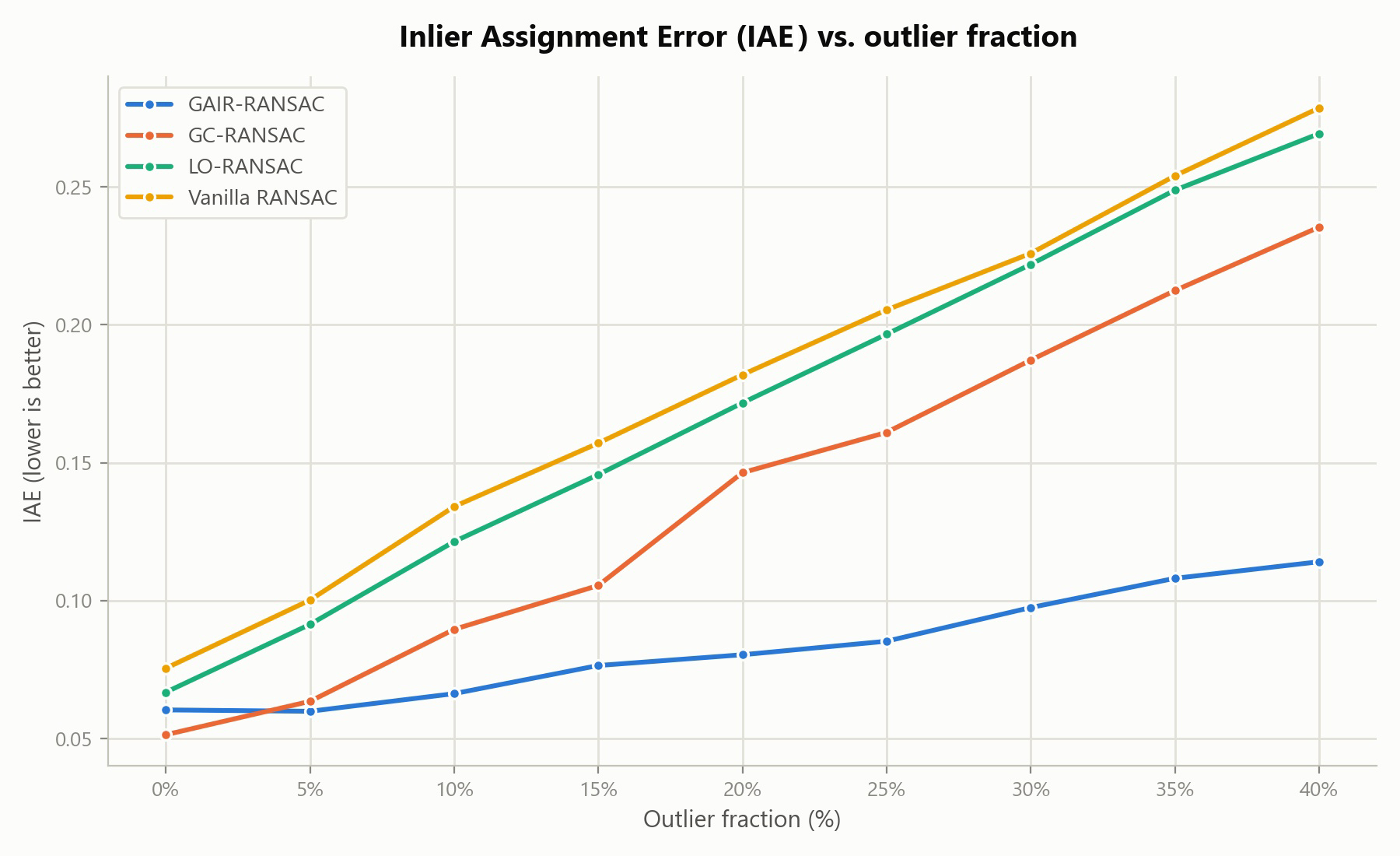}
        \caption{Inlier Assignment Error (IAE)}
        \label{fig:plotcd}
    \end{subfigure}
    
    \caption{
    Primitive decomposition results across datasets. 
    (a) and (b) report the average Chamfer Distance (CD) and Hausdorff Distance (HD) on \emph{SqSoup} and \emph{3D Shapes}. 
    (c) shows the Inlier Assignment Error (IAE) across both datasets. 
    GAIR-\ransacov consistently achieves lower error and better geometric accuracy.
    }
    \label{fig:multimodel_results}
\end{figure*}

We then extend the evaluation to all ten point clouds from \texttt{SqSoup} and \texttt{3DShapes}. For each algorithm, we run 25 trials of sequential \ransac and collect all candidate models generated across runs. These candidates are then passed to a \ransacov selection step, which selects the subset of at most $\kappa$ models that jointly maximizes the number of explained inliers. 
To better analyze the behavior of the different methods, we report the trend of the Inlier Assignment Error (IAE) as a function of the outlier ratio separately for the two datasets: \texttt{SqSoup} in Figure~\ref{fig:plotme1} and \emph{3D Shapes} in Figure~\ref{fig:plotme2}. This metric evaluates the consistency of point-to-model assignments without requiring explicit multi-label segmentation, making it applicable also to real-world shapes where ground-truth primitive annotations are not available (e.g., Thingi10K).
In both cases, GC-\ransacov and GAIR-\ransacov emerge as the best-performing methods, clearly outperforming Vanilla and LO-based formulations. This confirms, also in the multi-model setting, the benefit of moving beyond purely residual-based refinement strategies.

Averaging the results over all ten point clouds, reported in Table~\ref{tab:outlier_results}, confirms that  GAIR-\ransacov achieves the lowest \textcolor{black}{Chamfer Distance} at every outlier level, with a consistent margin over all baselines. This trend indicates that geometric-aware refinement is particularly effective in preserving reliable inlier assignments between multiple primitives even under severe contamination.

A qualitative example of this robustness is illustrated in Figure~\ref{fig:robustness}, where GAIR-\ransacov is applied to the \emph{Cartoon Character} shape under high outlier contamination. Despite the severe degradation of the input, the method still recovers a set of superquadrics that provides a coherent approximation of the overall geometry.

The same trend is also reflected by the \textcolor{black}{IAE reported in Table~\ref{tab:outlier_results}b} and in  Figure~\ref{fig:plotcd}. Except for the clean-data setting, where GC-RANSACOV obtains a slightly lower IAE, GAIR-RANSACOV achieves the lowest error at every non-zero outlier level. The gap becomes more pronounced as the contamination level increases, suggesting that geometric awareness becomes increasingly important when residual and proximity cues alone are no longer sufficient. \textcolor{black}{ In fact, at low outlier injection levels, all methods achieve similar IAE values despite exhibiting substantially different Chamfer distances.}

A qualitative example is shown in Figure~\ref{fig:bird}, where we report the primitive decomposition obtained by GAIR-\ransacov on the \emph{Bird} shape together with the point-wise Chamfer error with respect to the fitted superquadrics. As expected, the decomposition captures the overall structure of the shape while smoothing out fine-scale details. The largest errors are concentrated around high-curvature distinctive regions, such as the beak, which are less well approximated by a compact superquadric representation. These details could in principle be recovered by increasing the model complexity, e.g., by allowing a larger number of primitives.

\begingroup
\color{black}
\paragraph{Fixed run-time budget.}
To compare the methods under the same computational budget, we fix
the outlier fraction to $0.40$ and allow each method to generate
candidate models for $60$, $90$, or $120$ seconds. The numbers of
outer and inner iterations are fixed to $20$, and the resulting
candidates are passed to \ransacov, which selects the subset
maximizing inlier coverage.

As reported in Table~\ref{tab:budget_results_040},
GAIR-\ransacov achieves the lowest Chamfer Distance and Inlier
Assignment Error for every tested budget. Its performance improves
consistently as more computation becomes available, whereas the
competing approaches show smaller gains and tend to plateau.
Results are averaged over all ten point clouds of \textsl{SqSoup} and \textsl{3D Shapes} and $16$ independent
trials.

\begin{table}[t]
\centering
\caption{Chamfer distance (CD) and Inlier Assignment Error (IAE) as a function of the time budget, averaged fairly across point clouds (outlier fraction fixed at $0.40$).}
\label{tab:budget_results_040}

\textbf{Chamfer Distance $\downarrow$} \\[2pt]
\begin{tabular}{lccc}
\toprule
Algorithm & 60s & 90s & 120s \\
\midrule
GAIR-\ransacov    & \textbf{0.0987} & \textbf{0.0942} & \textbf{0.0921} \\
GC-\ransacov      & 0.1131          & 0.1101          & 0.1092          \\
LO-\ransacov      & 0.1064          & 0.1043          & 0.1039          \\
Vanilla \ransacov & 0.1077          & 0.1071          & 0.1069          \\
\bottomrule
\end{tabular}

\vspace{8pt}
\textbf{Inlier Assignment Error (IAE) $\downarrow$} \\[2pt]
\begin{tabular}{lccc}
\toprule
Algorithm & 60s & 90s & 120s \\
\midrule
GAIR-\ransacov    & \textbf{0.1880} & \textbf{0.1717} & \textbf{0.1661} \\
GC-\ransacov      & 0.2169          & 0.2092          & 0.2053          \\
LO-\ransacov      & 0.2000          & 0.1925          & 0.1886          \\
Vanilla \ransacov & 0.1959          & 0.1899          & 0.1869          \\
\bottomrule
\end{tabular}
\end{table}
\endgroup

\begingroup
\color{black}

\paragraph{Sketchfab point clouds.}
Additional qualitative comparisons on the publicly available
\textsl{Sketchfab scans} are reported in
Figure~\ref{fig:sketchfab}. For each example, we show the input
point cloud and the decompositions obtained with GC-\ransac
and GAIR-\ransacov. Both methods employ the same set-cover
model-selection stage and differ only in the inlier-refinement
strategy used during hypothesis generation. Therefore, the
differences observed in the final decompositions can be directly
attributed to the quality of the generated candidate models.

The proximity-based refinement of GC-\ransac tends to propagate
inlier assignments across adjacent but geometrically distinct surface
regions. As a result, the resulting hypothesis pool often contains
large primitives spanning multiple object parts, while lacking
candidates that separately represent the individual components.
The subsequent set-cover optimization cannot recover such missing
hypotheses and consequently produces under-segmented
decompositions. In contrast, the normal-aware refinement of GAIR
limits inlier propagation across surface discontinuities, generating
candidate primitives that better follow the visible geometric
components of the objects. This behavior is particularly evident for
the \textsl{hammer}, the \textsl{hydrant}, and the \textsl{baptismal font}, where GAIR-\ransacov
produces a decomposition that is more consistent with the main
geometric parts of the shapes, whereas GC-\ransacov
systematically merges distinct components.

\begin{figure*}[!t]
    \centering
    \begin{subfigure}[t]{0.15\textwidth}
        \centering
        \includegraphics[
            width=\linewidth,
            height=0.28\textheight,
            keepaspectratio
        ]{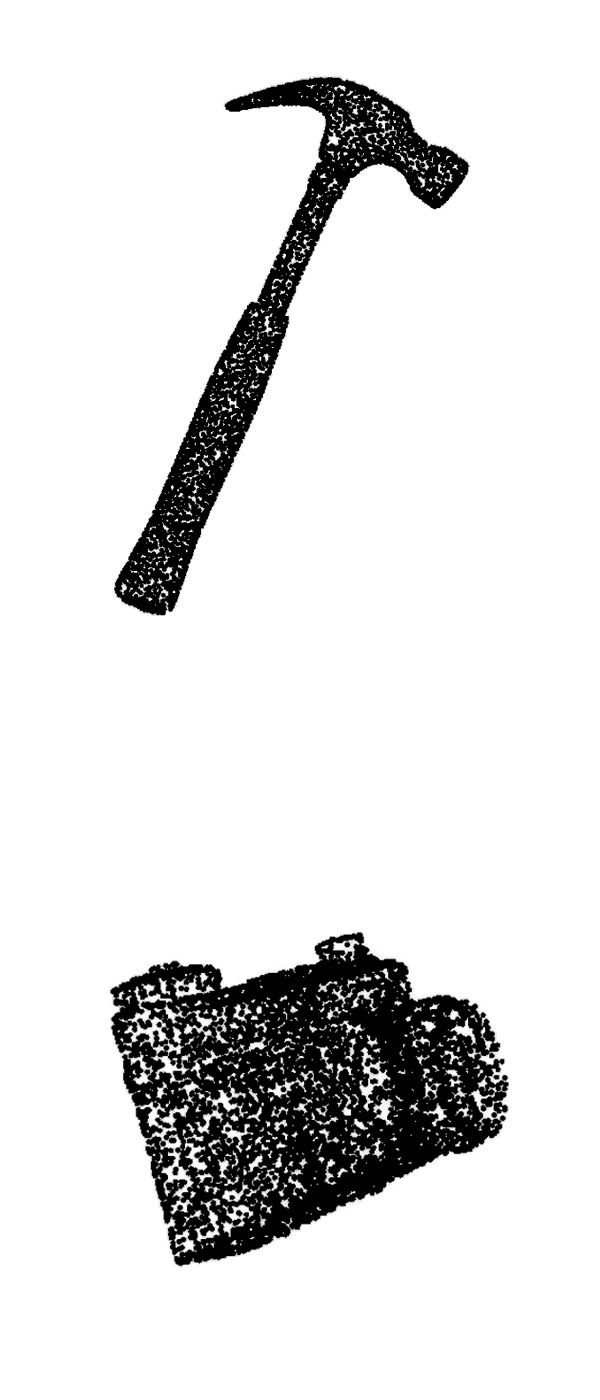}
        \caption{ \footnotesize Input point cloud}
    \end{subfigure}
    \hfill
    \begin{subfigure}[t]{0.15\textwidth}
        \centering
        \includegraphics[
            width=\linewidth,
            height=0.28\textheight,
            keepaspectratio
        ]{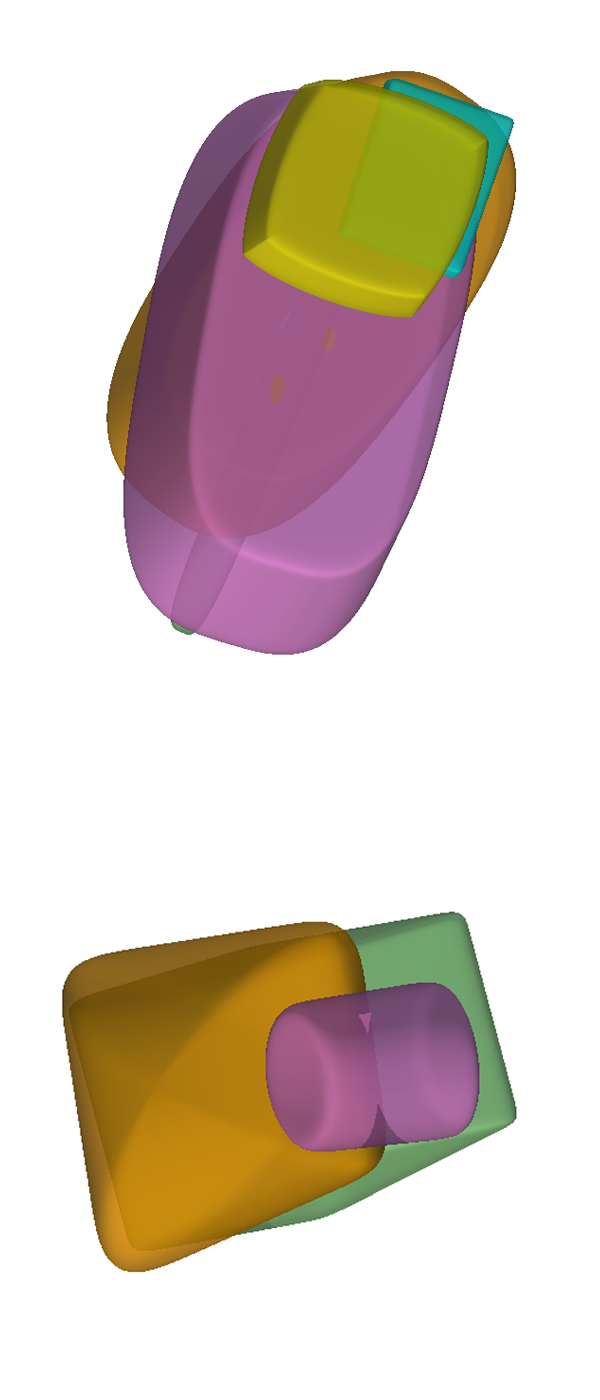}
        \caption{\footnotesize GC-\ransacov}
    \end{subfigure}
        \hfill
    \begin{subfigure}[t]{0.15\textwidth}
        \centering
        \includegraphics[
            width=\linewidth,
            height=0.28\textheight,
            keepaspectratio
        ]{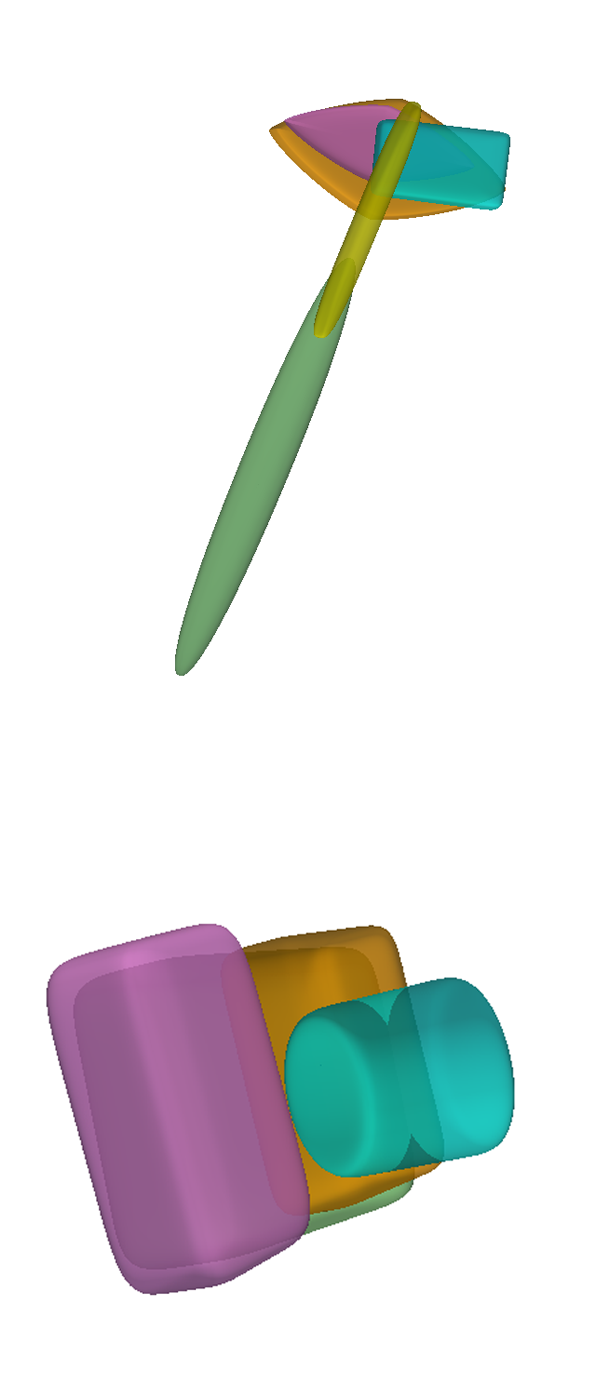}
        \caption{\footnotesize GAIR-\ransacov}
    \end{subfigure}
    \hspace{0.04\textwidth}
        \begin{subfigure}[t]{0.15\textwidth}
        \centering
        \includegraphics[
            width=\linewidth,
            height=0.28\textheight,
            keepaspectratio
        ]{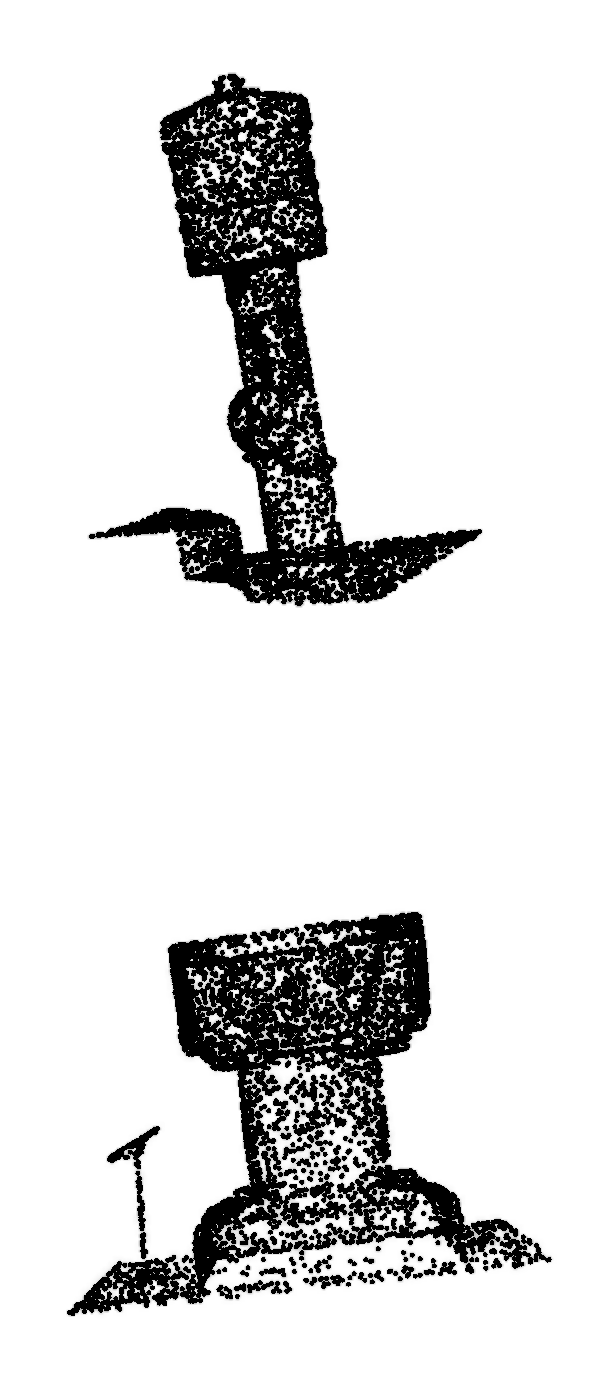}
        \caption{\footnotesize Input point cloud}
    \end{subfigure}
    \hfill
    \begin{subfigure}[t]{0.15\textwidth}
        \centering
        \includegraphics[
            width=\linewidth,
            height=0.28\textheight,
            keepaspectratio
        ]{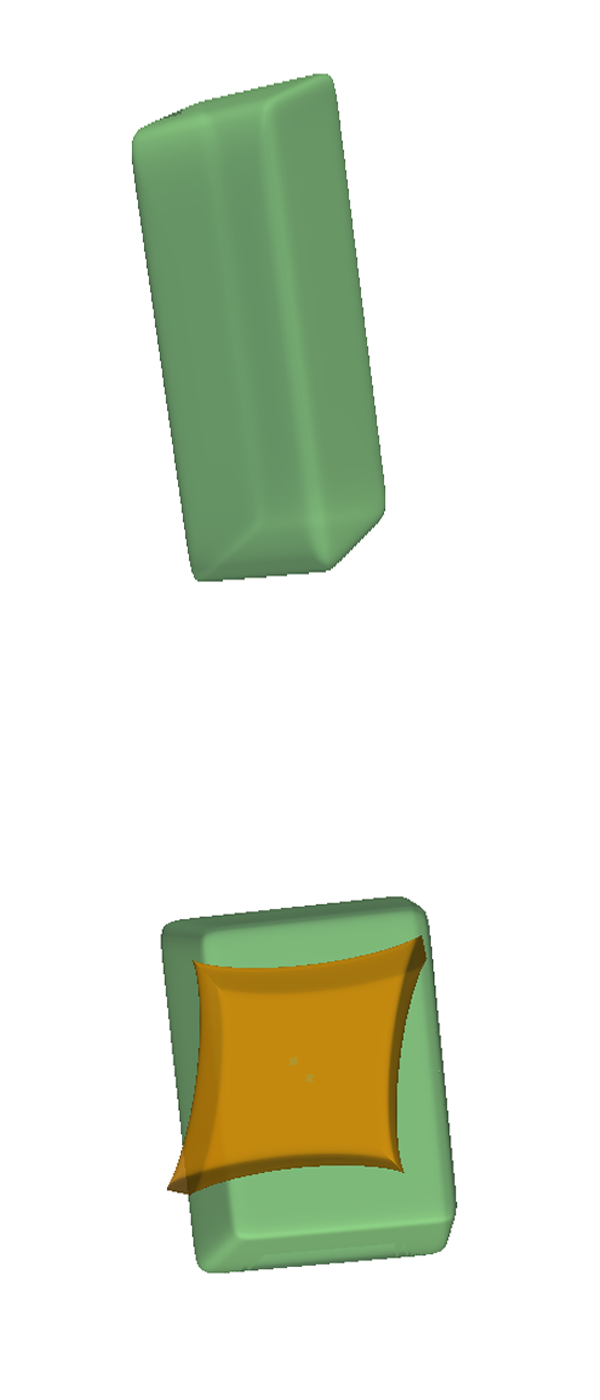}
        \caption{\footnotesize GC-\ransacov}
    \end{subfigure}
        \hfill
    \begin{subfigure}[t]{0.15\textwidth}
        \centering
        \includegraphics[
            width=\linewidth,
            height=0.28\textheight,
            keepaspectratio
        ]{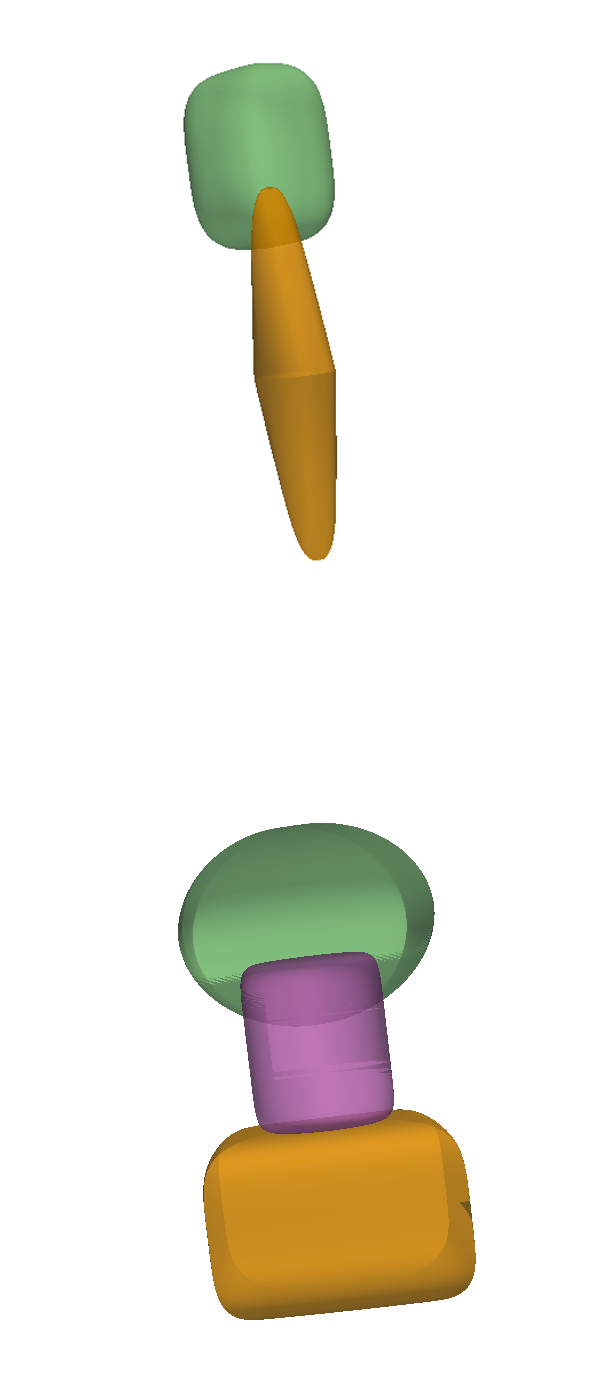}
        \caption{\footnotesize GAIR-\ransacov}
        \label{fig:energy_ablation_iae}
    \end{subfigure}
        \caption{Qualitative comparison on publicly available Sketchfab point
    clouds. Within each group, from left to right, we show the input
    point cloud, the decomposition obtained with GC-\ransacov, and
    the decomposition obtained with GAIR-\ransacov.}
    \label{fig:sketchfab}
\end{figure*}

\section{Ablation studies on the GAIR energy}

We perform a controlled ablation study on the \emph{Hammer}
point cloud, using eight independent runs for each configuration.
Recall that the refinement minimizes the energy
\begin{equation}
E(f)
=
\sum_{\vec{p}\in\mathcal D} E_1(f_p)
+
\sum_{(\vec{p},\vec{q})\in\mathcal E}
E_2(f_p,f_q),
\end{equation}
as defined in Eq.~\eqref{eq:energy}.

To clarify the contribution of the individual components, we denote
by $E_1^{\mathrm{GAIR}}$ the normal-aware unary term in
Eq.~\eqref{eq:unary}, by
$E_2^{\mathrm{GAIR}}(C)$ the residual-aware pairwise term in
Eq.~\eqref{eq:pairwise}, and by $E_2^{\mathrm{GC}}$ the original
GC-\ransac{} pairwise term in Eq.~\eqref{eq:gc_pairwise}.
We compare the following configurations:

\begin{enumerate}[noitemsep]

    \item \textbf{GC-\ransac}, using the original GC-\ransac{}
    unary and pairwise terms;

    \item \textbf{GAIR-U}, using the GAIR unary term
    $E_1^{\mathrm{GAIR}}$ together with the original pairwise term
    $E_2^{\mathrm{GC}}$. This variant isolates the effect of
    introducing point-to-model normal consistency in the unary cost;

    \item \textbf{GAIR-Uniform}, using
    $E_1^{\mathrm{GAIR}}$ and the GAIR pairwise formulation
    $E_2^{\mathrm{GAIR}}(C)$, but setting
    \[
        C(\vec{n}_p,\vec{n}_q)\equiv 1
    \]
    for every neighboring pair. This removes normal coherence from
    the pairwise term while retaining its residual-aware structure;

    \item \textbf{GAIR-Hard}, using the same GAIR unary and pairwise
    terms, but replacing the continuous coherence with the binary
    function
    \[
        C_{\mathrm{hard}}(\vec{n}_p,\vec{n}_q)
        =
        \mathbb{1}\!\left[
        C(\vec{n}_p,\vec{n}_q)>0.9
        \right].
    \]
    Normal information therefore determines whether an edge is
    active, but all retained edges receive the same weight;

    \item \textbf{GAIR}, using the complete formulation with
    $E_1^{\mathrm{GAIR}}$ and
    $E_2^{\mathrm{GAIR}}(C)$, including hard edge removal below the
    coherence threshold and continuous coherence weighting on the
    retained edges.

\end{enumerate}

\begin{figure*}[htb]
    \centering

    \begin{subfigure}[t]{0.48\textwidth}
        \centering
        \includegraphics[
            width=\linewidth,
            height=0.28\textheight,
            keepaspectratio
        ]{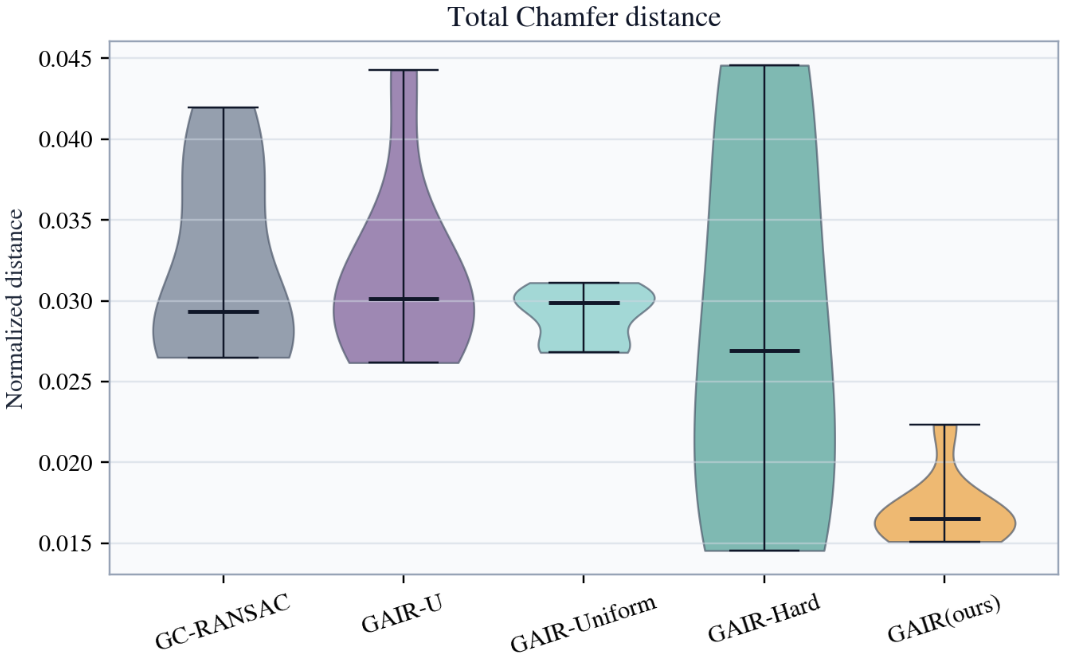}
        \caption{Chamfer Distance obtained with different energy formulations. Lower values are better.}
        \label{fig:energy_ablation_chamfer}
    \end{subfigure}
    \hfill
    \begin{subfigure}[t]{0.48\textwidth}
        \centering
        \includegraphics[
            width=\linewidth,
            height=0.28\textheight,
            keepaspectratio
        ]{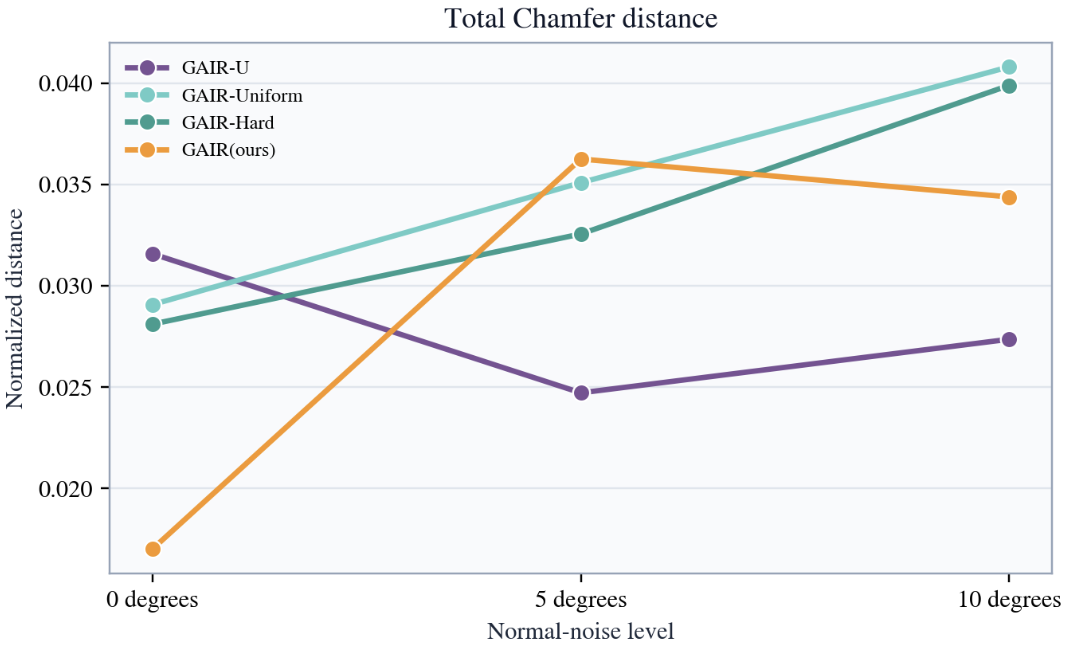}
        \caption{Impact of the noise on normals on the energy formulations. Lower values are better}
        \label{fig:energy_ablation_normal_noise}
    \end{subfigure}

    \caption{Ablation studies on the energy terms, tested on the \textsl{Hammer} point cloud.}
    \label{fig:energy_ablation}
\end{figure*}

Figure~\ref{fig:energy_ablation_chamfer} shows that
GC-\ransac{} and GAIR-U achieve comparable results. Thus, replacing
only the unary term with its normal-aware counterpart is not
sufficient to substantially improve the reconstruction.

GAIR-Uniform achieves a similar average error but reduces the
variability across runs. Since this formulation does not use normal
coherence in the pairwise term, the improvement can be attributed
to the residual-aware structure of
$E_2^{\mathrm{GAIR}}$, which provides a more stable regularization
than the original GC-\ransac{} pairwise term.

Introducing hard normal-based edge selection in GAIR-Hard lowers
the average error, confirming that preventing propagation across
normal discontinuities is important. However, this formulation also
exhibits substantially higher variability. The binary coherence
introduces a discontinuous decision at the threshold, and all retained
edges receive the same weight, including marginal connections whose
coherence is only slightly above $0.9$. Small variations in the
estimated normals or in the sampled hypotheses can therefore produce
markedly different label propagations.

The complete GAIR formulation retains the benefit of removing
incompatible edges while weighting the remaining connections
according to their actual normal coherence. This reduces the
influence of uncertain connections and yields both the lowest average
error and the most stable results.

\paragraph{Sensitivity to normal perturbations.}
Since GAIR relies on surface normals, we additionally evaluate its
sensitivity to normal-estimation errors. Each input normal is
independently perturbed with zero-mean Gaussian noise and
renormalized. The perturbation magnitude is selected to produce
different mean angular deviations, including approximately
$5^\circ$ and $10^\circ$. As shown in Figure~\ref{fig:energy_ablation_normal_noise}, the
advantage of the normal-aware formulations decreases as the angular
error increases. In particular, the variants using normal coherence in
the pairwise term are the most affected, since corrupted normals alter
both the graph connectivity and the edge weights. When the normals
become severely inaccurate, formulations that do not rely on them are
preferable, as expected.

\section{Primitive decomposition of real LiDAR scans}
\label{sec:real_scans}

\begin{figure*}[t]
    \centering

    \begin{minipage}[t]{0.035\linewidth}
        \vspace{0pt}

        \begin{minipage}[c][0.08\textheight][c]{\linewidth}
            \centering
            \rotatebox[origin=c]{90}{%
                \footnotesize\normalfont Ball and box%
            }
        \end{minipage}

        \begin{minipage}[c][0.06\textheight][c]{\linewidth}
            \centering
            \rotatebox[origin=c]{90}{%
                \footnotesize\normalfont Sofa%
            }
        \end{minipage}

        \begin{minipage}[c][0.15\textheight][c]{\linewidth}
            \centering
            \rotatebox[origin=c]{90}{%
                \footnotesize\normalfont Plush toy%
            }
        \end{minipage}
    \end{minipage}%
    \hspace{0.002\linewidth}%
    \begin{subfigure}[t]{0.285\linewidth}
        \vspace{0pt}
        \centering
        \includegraphics[
            width=\linewidth,
            height=0.28\textheight,
            keepaspectratio
        ]{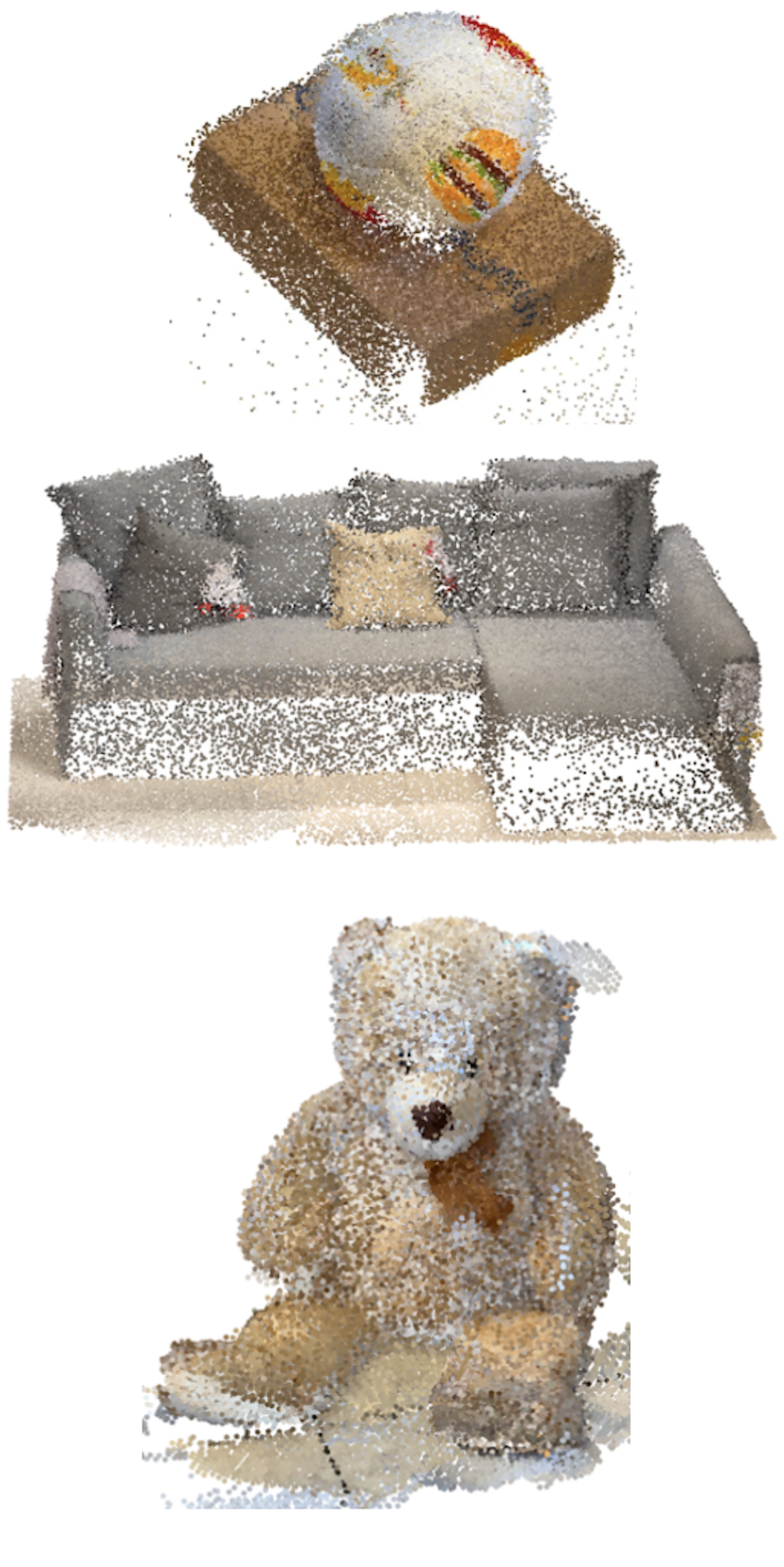}
        \caption{Input point cloud.}
        \label{fig:real_pc_input}
    \end{subfigure}%
    \hfill
    \begin{subfigure}[t]{0.285\linewidth}
        \vspace{0pt}
        \centering
        \includegraphics[
            width=\linewidth,
            height=0.28\textheight,
            keepaspectratio
        ]{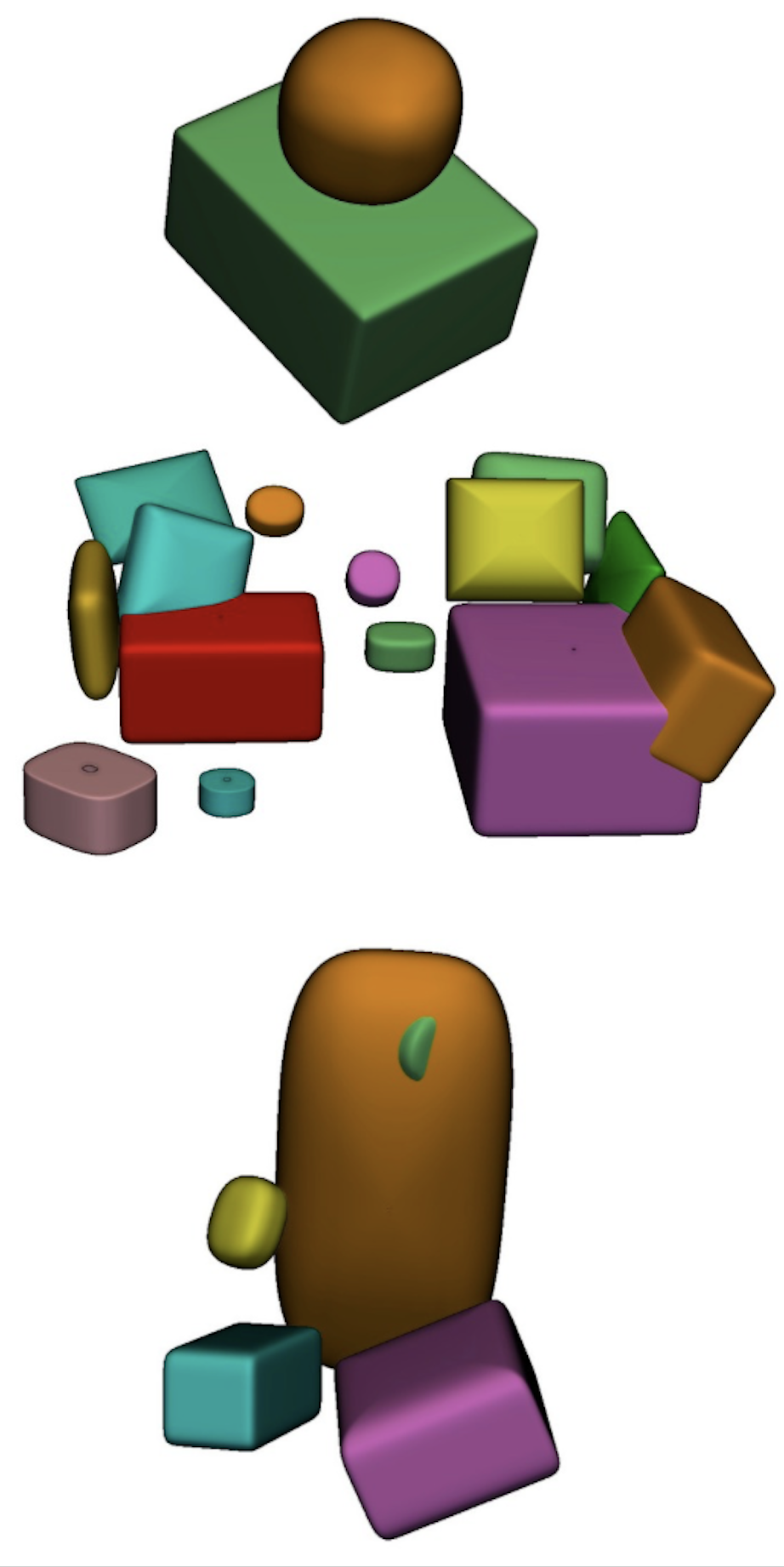}
        \caption{GC-\ransac decomposition.}
        \label{fig:real_pc_gc}
    \end{subfigure}%
    \hfill
    \begin{subfigure}[t]{0.285\linewidth}
        \vspace{0pt}
        \centering
        \includegraphics[
            width=\linewidth,
            height=0.28\textheight,
            keepaspectratio
        ]{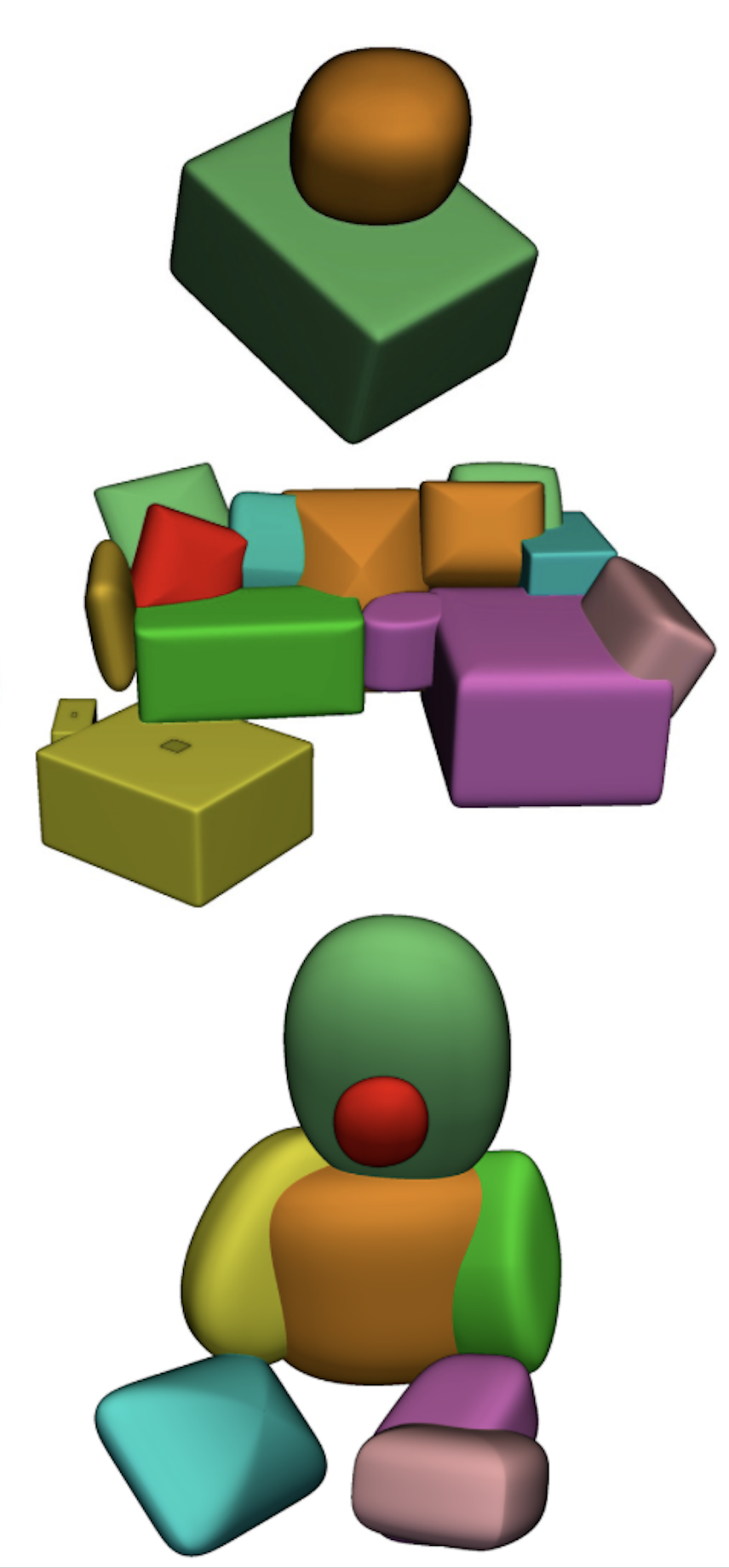}
        \caption{GAIR-\ransac decomposition.}
        \label{fig:real_pc_gair}
    \end{subfigure}

    \caption{Qualitative comparison on 3D real scanned point
    clouds captured with an iPhone 16 Pro equipped with LiDAR. Within each
    group, from left to right, we show the input point cloud, the
    decomposition obtained with GC-\ransac, and the decomposition obtained
    with GAIR-\ransac.}
    \label{fig:real_pc}
\end{figure*}

We evaluate the algorithms on three real point-cloud scenes exhibiting realistic acquisition artifacts, including non-uniform sampling, missing regions, and imperfect surface coverage. In particular, object surfaces in
contact with the floor are generally not visible. Surface normals are
estimated directly from the noisy acquired geometry.

Since all scans are represented in metric scale, we fix the inlier
threshold to $\varepsilon=0.015\,\mathrm{m}$ and keep it unchanged
across all scenes. For particularly dense scans, we apply uniform
subsampling before decomposition. In particular, the ball and box and the sofa scans are
reduced to $100$k points, preserving their main geometric structure
while substantially reducing the computational cost. The Plush  toy dataset is less dense and has 56k points.

\paragraph{Ball and box.}
The first scene consists of a ball placed on top of a box. Although
the acquisition contains realistic occlusions and the lower face of
the box is not observed, the underlying geometry is simple and
consists of only two clearly distinguishable primitives. Consequently,
both GC-\ransac and GAIR-\ransac recover the expected
decomposition, showing that both methods behave correctly in a
geometrically simple real-world setting.

\paragraph{Sofa.}
The second scene contains a sofa with several cushions. This
configuration is substantially more challenging, since it contains
many adjacent primitive-like components, several contact regions,
and cushions whose surfaces are only partially visible. In this
setting, proximity alone is not sufficient to distinguish neighboring
parts: the refinement of GC-\ransac tends to propagate inliers
across contact regions and fails to recover some of the more sparsely
sampled lower components. In contrast, the normal-aware refinement
of GAIR-\ransac better preserves the boundaries between
adjacent cushions and produces a more complete decomposition of
the scene.

\paragraph{Plush toy.}
The plush toy represents the most challenging example. Unlike the
sofa components, its body parts are only approximately represented
by superquadrics. Moreover, its fuzzy surface, weakly separated
components, and multiple contact regions result in noisy normals and
poorly defined geometric boundaries. GC-\ransac produces a
strongly under-segmented solution, using a single large superquadric
to approximate both the head and torso, while the remaining
primitives provide an irregular approximation of the legs, arms, and
muzzle. GAIR-\ransac instead obtains a decomposition that is
more consistent with the visible object structure, separately
representing the head, torso, limbs, and muzzle. Some fine details,
such as one foot and part of one limb, are nevertheless not recovered.

Overall, these experiments show that GAIR transfers effectively to
noisy and incomplete LiDAR acquisitions and improves the
decomposition of objects composed of several adjacent parts.
Nevertheless, fine part-level decomposition remains difficult when
the input scan provides insufficient surface coverage or lacks clear
geometric separation between components.

\endgroup

\section{ Future Work and Conclusions}

Several promising directions remain open for future investigation:
\textbf{Additional geometric priors.} The energy formulation can be naturally extended to incorporate other geometric constraints. An exciting direction would be to integrate a term based on Chamfer distance, which could further regularize the solution by promoting the selection of compact superquadrics that provide better coverage of the point data while discouraging excessive surface extensions into empty regions.

\textbf{Integration with global multi-model fitting.} An important direction is to explore how the proposed inlier refinement can be embedded within more sophisticated multi-model fitting frameworks. A promising approach would reformulate the problem as a global multi-label optimization, where each label corresponds to a different model (plus one for outliers). This would reduce the dependence on the order of model extraction and potentially lead to more robust decompositions in complex scenes.

\textbf{Automatic model complexity selection.} Finally, integrating a model complexity term directly into the \ransac loop could automatically determine the number of primitives $\kappa$, moving beyond the current requirement of specifying it in advance. This would bring the method closer to a fully automatic primitive decomposition pipeline.

This work presents an algorithmic contribution to the problem of primitive decomposition in 3D point clouds. At its core lies a novel energy formulation that extends the GC-\ransac framework by incorporating geometric priors specifically tailored for primitive fitting. While we focus primarily on normal consistency in this work, the proposed formulation is general and provides a sound foundation for integrating additional geometric constraints.
We demonstrate how this formulation can be effectively integrated into both single-model and multi-model estimation pipelines, leading to the GAIR-\ransac and GAIR-\ransacov frameworks. The former improves the stability and accuracy of superquadric fitting in the presence of noise and outliers, while the latter enables robust primitive decomposition in challenging scenarios with adjacent or overlapping structures.
Although graph-cut optimization is computationally more expensive per iteration than the refinement step in LO-\ransac, our experiments reveal a favourable tradeoff: GAIR remains faster than GC-\ransac in our experiments while consistently providing more accurate decompositions. This results in both better accuracy and more consistent execution times overall.

\section*{Acknowledgments}
This work is supported by GEOPRIDE under ID 202224 5ZYB and CUP D53D23008370001 (PRIN 2022 M4.C2.1.1 Investment).

\appendix

\printcredits
\bibliographystyle{cas-model2-names}

\bibliography{bibliography}

\end{document}